\documentclass[10pt,twocolumn,letterpaper]{article}

\usepackage[pagenumbers]{cvpr} %

\usepackage[dvipsnames]{xcolor}

\definecolor{cvprblue}{rgb}{0.21,0.49,0.74}
\usepackage[pagebackref,breaklinks,colorlinks,citecolor=cvprblue]{hyperref}
\usepackage{xcolor}
\usepackage{amsmath}
\usepackage{amssymb}
\usepackage{amsfonts}
\usepackage{dsfont}
\usepackage{array}
\usepackage{multirow}
\usepackage{stfloats}
\usepackage[dvipsnames]{xcolor}

\usepackage{algorithm}
\usepackage{algpseudocode}

\title{CONFORM: \underline{Con}trast is All You Need \underline{For} High-Fidelity Text-to-Image Diffusion \underline{M}odels}

\author{Tuna Han Salih Meral$^{1}$ \quad Enis Simsar$^{2}$ \quad Federico Tombari$^{3}$ \quad Pınar Yanardağ$^{1}$ \\[2mm]
$^{1}$Virginia Tech \qquad  $^{2}$ETH Zürich-DALAB \qquad $^{3}$Google\\
{\tt\small \href{https://conform-diffusion.github.io}{https://conform-diffusion.github.io}}
}

\begin{document}

\twocolumn[{
    \maketitle
    \vspace{-12pt}
    \includegraphics[width=\linewidth]{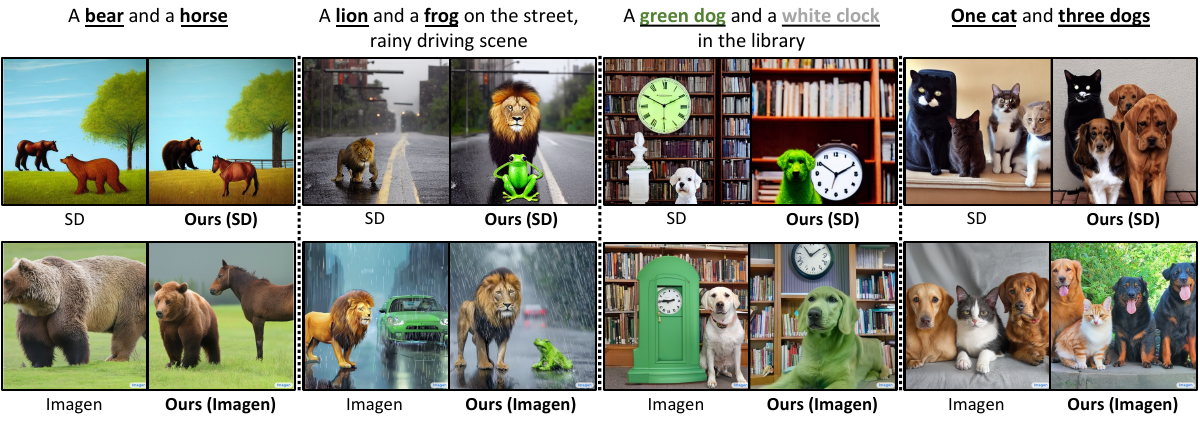}
    \captionof{figure}{Our training-free method combines a contrastive objective with test-time optimization, significantly improving how models such as Imagen and Stable Diffusion generate images with text prompts consisting of multiple concepts or subjects such as \textit{`a \underline{\textbf{bear}} and a \underline{\textbf{horse}}'}.}\vspace{0.3cm}
    \label{fig:teaser}
}]

{
    \renewcommand{\thefootnote}{\fnsymbol{footnote}}
    \footnotetext{Contact author: Tuna Han Salih Meral (\textit{tmeral@vt.edu})}
}

\maketitle

\begin{abstract}
Images produced by text-to-image diffusion models might not always faithfully represent the semantic intent of the provided text prompt, where the model might overlook or entirely fail to produce certain objects. Existing solutions often require customly tailored functions for each of these problems, leading to sub-optimal results, especially for complex prompts. Our work introduces a novel perspective by tackling this challenge in a contrastive context. Our approach intuitively promotes the segregation of objects in attention maps while also maintaining that pairs of related attributes are kept close to each other.  We conduct extensive experiments across a wide variety of scenarios, each involving unique combinations of objects, attributes, and scenes. These experiments effectively showcase the versatility, efficiency, and flexibility of our method in working with both latent and pixel-based diffusion models, including Stable Diffusion and Imagen. Moreover, we publicly share our source code to facilitate further research.
\end{abstract}    
\section{Introduction}
\label{sec:intro}

\begin{figure}[t!]
    \centering
    \includegraphics[width=1\linewidth]{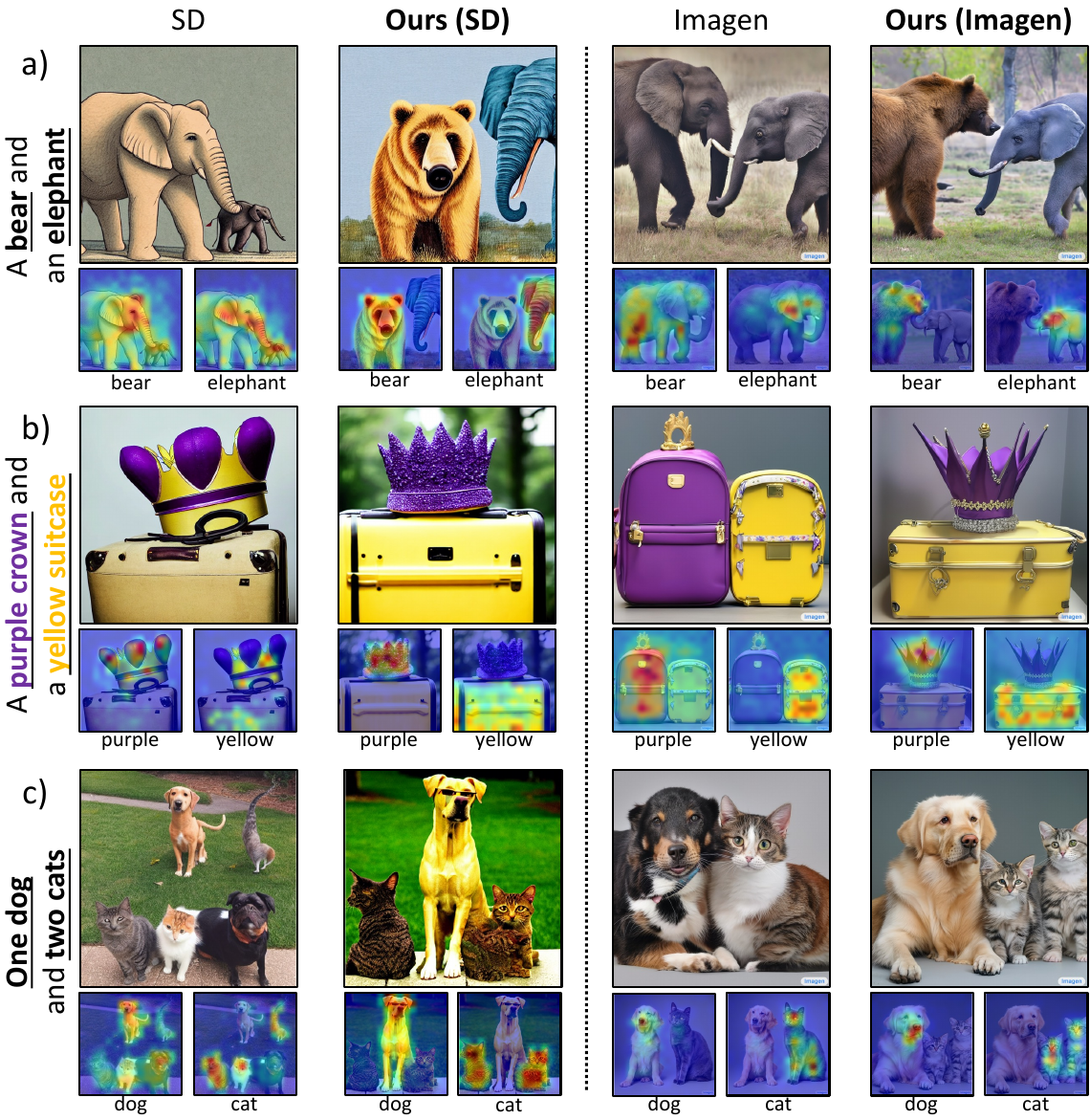}
    \caption{\textbf{Failure cases of Stable Diffusion \cite{stable-diffusion} and Imagen \cite{imagen}.} Text-to-image diffusion models may not faithfully adhere to the subjects specified in the text prompt: a) missing objects (\eg{,} \textit{bear}), b) misaligned attributes (\eg{,} \textit{the color yellow blends into the crown}), and c) inaccurate object count (\eg{,} \textit{only one cat is generated instead of two}). Our method steers the diffusion process towards more faithful images in both SD and Imagen.}
    \label{fig:failure}
    \vspace{-1em}
\end{figure}

Diffusion text-to-image models \cite{ddpm} have showcased remarkable progress in generating images using textual cues \cite{imagen, stable-diffusion, dalle2}. These models offer a wide set of capabilities, ranging from image editing \cite{blended,blended_latent,diffedit,pix2pix-zero,pnp,p2p}, personalized content creation~\cite{ruiz2022dreambooth}, and inpainting \cite{lugmayr2022repaint}. However, images produced by these models might not always faithfully represent the semantic intent of the given text prompt \cite{chefer2023attend, tang2022daam}. Notable semantic discrepancies in models like Stable Diffusion \cite{stable-diffusion} and Imagen \cite{imagen} include a) \textit{missing objects} where the model might overlook or entirely fail to produce certain objects; b) \textit{attribute binding} where the model might mistakenly link attributes to the wrong subjects \cite{chefer2023attend}; and c) \textit{miscounting} where the model fails to accurately produce the right quantity of objects \cite{li2023divide, yu2022parti}. \Cref{fig:failure} illustrates these shortcomings in popular diffusion models, Stable Diffusion \cite{stable-diffusion} and Imagen \cite{imagen}. For example, the output might neglect certain subjects, as in the `a bear and an elephant' prompt, where the bear is ignored as depicted in Fig.~\ref{fig:failure}(a). Additionally, the model might mix up attributes, such as mixing the colors in the `a purple crown and a yellow suitcase' prompt as seen in Fig.~\ref{fig:failure}(b). Another behavior that is often attributed to the imprecise language comprehension of the CLIP text encoder \cite{radford2021learning, paiss2023teaching} is the failure to produce the correct quantity of subjects as in Fig.~\ref{fig:failure}(c) where the model either produces an excessive number of cats (SD or failed to include a cat (Imagen) for `one dog and two cats' prompt.

\begin{figure}[t!]
    \centering
    \includegraphics[width=1\linewidth]{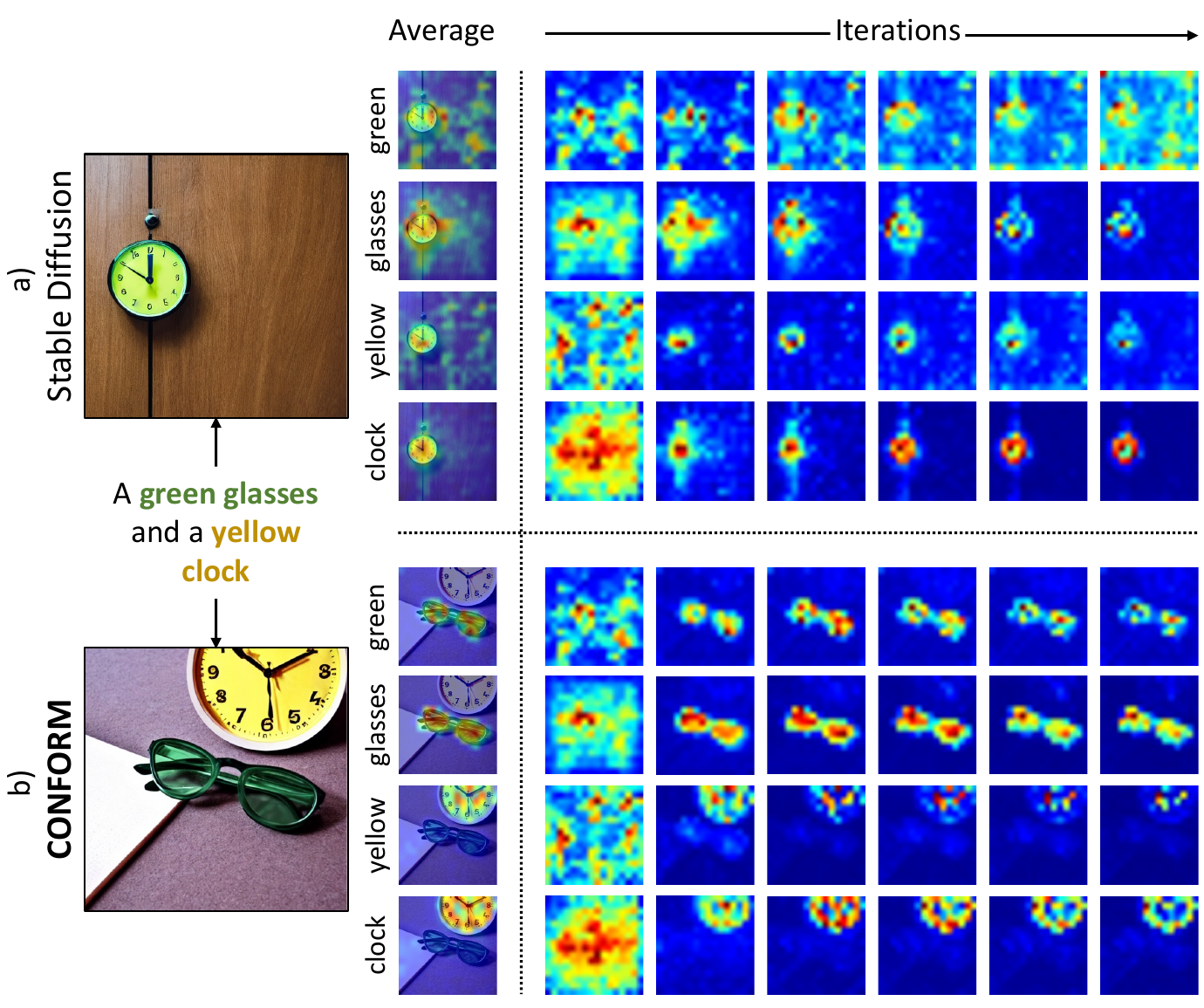}
    \caption{\textbf{Attention scattering in backward process.} In Stable Diffusion, the attention to attributes like \textit{green} and \textit{yellow} dissolves over backward timesteps (a). Our method effectively preserves these attention maps (b).}
    \label{fig:timestep-failure}
    \vspace{-1.5em}
\end{figure}
  
Recent studies proposed various solutions to these semantic challenges \cite{chefer2023attend, li2023divide, agarwal2023star, wu2023harnessing, kim2023dense}. For example, Chefer et al. \cite{chefer2023attend} optimize cross-attention maps to encourage object presence, while Li et al. \cite{li2023divide} use a dual loss function to segregate the attention map into distinct areas of attention and to reinforce attribute association. Kim et al. \cite{kim2023dense} enhance fidelity by directly adjusting intermediate attention maps according to user-specified layouts. However, a common limitation of these methods is their reliance on tailored objective functions for each issue, leading to sub-optimal performance or challenges when dealing with complex prompts.

 Attention maps, which depict the relationship between the input text and the generated pixels, offer a valuable lens for understanding these challenges, as emphasized by prior research \cite{chefer2023attend, agarwal2023star, p2p}.  For example, for the `a bear and an elephant' prompt, a significant overlap is observed in the cross-attention maps dedicated to each subject (refer to Fig.~\ref{fig:failure}(a)). This overlap makes it difficult to differentiate the two subjects and leads the model to produce only \textit{elephants}. Similarly, when prompted to produce a \textit{purple crown} and a \textit{yellow suitcase}, the attentions designated for \textit{purple} and \textit{yellow} are misaligned, causing the model to mistakenly mix colors of both (see the Fig.~\ref{fig:failure}(b)). Regarding counting, the attention maps usually concentrate solely on one region (refer to Fig.~\ref{fig:failure}(c) Imagen), resulting in the generation of an incorrect number of objects, such as \textit{one} cat instead of \textit{two}. Additionally, during the backward process, the attention maps corresponding to various attributes tend to scatter (See Fig.~\ref{fig:timestep-failure}). Therefore, to effectively reduce the scattering and ensure more focused and coherent attention allocation, we incorporated attention maps from the previous iteration. This enhances the model's ability to maintain consistency across the generation process, as seen in Fig.~\ref{fig:timestep-failure}.

In this work, we tackle the challenge of high-fidelity generation in text-to-image models within a contrastive framework. This framework considers the attributes of a specific object as positive pairs while contrasting them against attributes and objects outside their pairing. For example, in the prompt \textit{‘a green dog and a white clock’} (see Fig.~\ref{fig:teaser}), \textit{green} and \textit{dog} are treated as mutual positives, while \textit{white} and \textit{clock} become their contrastive counterparts, and vice versa. This approach separates distinct objects within the attention map, addressing the overlapping attention, and encourages distinct high-response areas for each object and attribute.  As a result, objects are distinctly separated from one another while being closely associated with their specific attributes, ensuring that the attention map represents both concepts effectively (see Fig.~\ref{fig:teaser} and \ref{fig:failure}).

The key contributions of our work are as follows:

\begin{itemize}
    \item We propose a training-free method utilizing a contrastive objective combined with test-time optimization to enhance the fidelity of pre-trained text-to-image diffusion models.
    \item Our approach is model-agnostic, applicable to popular text-to-image diffusion models like Stable Diffusion and Imagen.
    \item Our comprehensive experiments demonstrate the superiority of our method over baselines and competing approaches, evidenced by its performance on various benchmark datasets and user studies.
\end{itemize}

\section{Related work}
\label{sec:related}

\paragraph{Text-to-image diffusion models.} Before diffusion-based large-scale conditional image generation models, generative adversarial networks \cite{ye2021improving, tao2022df, zhu2019dm, zhang2021cross, xu2018attngan, kang2023gigagan}, variational autoencoders \cite{huang2017real}, and autoregressive models \cite{ramesh2021zero, yu2022scaling} were the main focus for both conditional and unconditional image synthesis. However, with the advent of diffusion-based image generation models \cite{ddpm, ddim, nichol2021glide}, and their evolution into large-scale text-to-image models \cite{balaji2022ediffi, imagen, dalle2}, they became the state-of-the-art for the text-to-image generation. Although the quality of generated samples increased significantly, it is still a challenge to create images that are faithful to the input prompt. Classifier-free guidance \cite{ho2022classifier} is introduced to enhance text reliance but there is still a need for prompt engineering \cite{liu2022design, wang2022diffusiondb, witteveen2022investigating} to produce input prompts so that the generated samples satisfy the intended properties specified in the input prompts.

\vspace{-12pt}

\paragraph{Improving the fidelity of text-to-image diffusion models.}

The challenge of aligning text-to-image model outputs with input prompts has been discussed in \cite{tang2022daam}. They identified that adjectival modifiers and co-hyponyms result in entangled features in cross-attention maps. To address this, \citet{liu2022compositional} introduced ComposableDiffusion allowing users to apply conjunction and negation operators in prompts to guide concept composition. Similarly, StructureDiffusion \cite{feng2023structureDiffusion} segments the prompts into noun phrases for more precise attention distribution. Wu et al. \cite{wu2023harnessing} developed an algorithm with a layout predictor for spatial layout generation, addressing the cross-attention map control. Agarwal et al. \cite{agarwal2023star} proposed A-star to minimize concept overlap and change in attention maps through iterations. Kim et al. \cite{kim2023dense} proposed DenseDiffusion, for region-specific textual feature accumulation. Chefer et al. \cite{chefer2023attend} focused on enhancing attention to neglected tokens, and \citet{li2023divide} proposed two separate tailored objective functions to address the missing objects and wrong attribute binding problems separately. Although these methods are taking steps forward to resolve the mentioned issues, they fail in several cases (see \cref{fig:main_comparison}).  The Attend and Excite method addresses solely the issue of neglected objects, but it falls short in effectively resolving the problem when the areas of maximum attention are close. On the other hand, Divide and Bind provides an approach to tackle the issue of incorrect attribute binding. However, its use in situations where the tokenizer of text embedding divides single object words into multiple tokens is unclear.

\section{Methodology}
\label{sec:methodology}

\begin{figure}[t]
    \centering
    \includegraphics[width=0.8\linewidth]{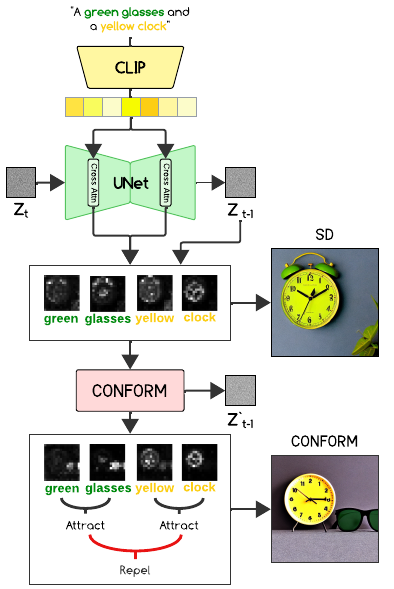}
    \caption{\textbf{An overview of CONFORM}. Given a prompt (\eg{,} \textit{`A green glasses and a yellow clock'}), we extract the subject tokens \textcolor{ForestGreen}{\textit{green}}, \textcolor{ForestGreen}{\textit{glasses}}, \textcolor{Dandelion}{\textit{yellow}}, and \textcolor{Dandelion}{\textit{clock}} and their corresponding attention maps ($A^{\texttt{green}}, A^{\texttt{glasses}}, A^{\texttt{yellow}}, A^{\texttt{clock}}$) from timesteps $t$ and $t+1$. We employ our contrastive objective at each time step to repel negative pairs and attract positive pairs.}
    \label{fig:framework}
    \vspace{-1em}
\end{figure}

In this section, we begin by outlining the basics of diffusion models and contrastive learning, followed by a detailed discussion of our methodology. An overview of our method is shown in Fig.~\ref{fig:framework}.

\subsection{Diffusion models}
\label{sec:background}

We applied our novel approach to two leading text-to-image models: Stable Diffusion (SD) and Imagen. Stable Diffusion operates in the latent space of an autoencoder, where an encoder $\mathcal{E}$ converts the input image $x$ into a lower-dimensional latent code $z = \mathcal{E}(x)$. The decoder $\mathcal{D}$ then reconstructs this latent back into the image space, achieving  $\mathcal{D}(z) \approx x$. On the other hand, Imagen operates within pixel space, extending its output via two consecutive image-to-image diffusion models for super-resolution.

Upon having a trained autoencoder, Stable Diffusion employs a diffusion model \cite{ddpm} that is trained within the latent space of the autoencoder. The training process involves gradually adding noise to the original latent code $z_0$ over time, leading to the generation of $z_t$ at timestep $t$. This latent code $z_0$ is in pixel space in Imagen and latent space in Stable Diffusion. A UNet \cite{unet} denoiser, denoted as $\epsilon_{\theta}$, is trained to predict the noise added to $z_0$. The training objective is formally expressed as:
\begin{equation}
    \mathcal{L} = \mathbb{E}_{z_t, \epsilon \sim \mathrm{N(0, I)}, c(\mathcal{P}), t } \left[ \Vert \epsilon - \epsilon_{\theta}(z_t, c(\mathcal{P}), t )\Vert ^ {2} \right]
\end{equation}
where $c(\mathcal{P})$ represents the conditional information and $\mathcal{P}$ is the text prompt fed to the text embedding model.

In Stable Diffusion, the sequential embedding of CLIP \cite{radford2021learning} model $c$ is supplied to a UNet network through a cross-attention mechanism, serving as conditioning to generate images that adhere to the provided text prompt $\mathcal{P}$. In Imagen, a pre-trained T5 XL language model \cite{raffel2020exploring} is used as a text-encoder instead. The cross-attention layers perform a linear projection of $c$ into queries ($Q$) and values ($V$), and they map intermediate representations from UNet to keys ($K$). Then, the attention at time $t$ is calculated as $A_t = \mathrm{Softmax}(QK^\intercal / \sqrt{d})$.
Notably the attention map at timestep $t$, $A_t$, can be reshaped into $\mathbb{R}^{h \times w \times l}$, where $h$, $w$ represents the resolution of the feature map, which can take values from $\{16\times16, 32\times32, 64\times64\}$, and $l$ corresponds to the sequence length of the text embedding. In our work, we primarily focus on the $\{16\times16\}$ attention maps, as they have been identified by Hertz et al. \cite{p2p} as the most semantically meaningful attention maps.
\subsection{Contrastive learning}

Contrastive learning has recently gained substantial popularity, delivering state-of-the-art results across multiple unsupervised representation learning tasks \cite{chen2020simple, oord2018representation, tian2019contrastive, hadsell2006dimensionality, chopra2005learning}. The core objective of contrastive learning is to develop representations that bring similar data points closer while pushing dissimilar data points apart. Let $x \in \mathcal{X}$ represent an input data point. We can define $x^{+}$ as a positive pair, where both data points, $x$ and $x^{+}$, share the same label, and $x^{-}$ as a negative pair, in which the data points have different labels. The kernel $f: \mathcal{X} \rightarrow \mathbb{R}^N$, takes an input $x$ and generates an embedding vector. InfoNCE, also known as NT-Xent, \cite{chen2020simple, he2020momentum, oord2018representation} is one of the popular contrastive learning objectives defined as follows:

\begin{equation}
    \mathcal{L} = -\log \frac{\exp(f(x) \cdot f(x^{+})/\tau)}{\sum_{i=0}^{M} \exp(f(x)\cdot f(x_i)/\tau)}
\end{equation}

In this equation, $\tau$ is the temperature parameter, regulating the penalties. The summation is performed over one positive sample, denoted as $x^{+}$, and $M$ negative samples. Essentially, this loss can be interpreted as the log loss of a softmax-based classifier aiming to classify the data point $x$ as the positive sample $x^{+}$. We utilized InfoNCE loss since we will operate on very limited data and need an objective function supporting fast convergence.

\begin{figure*}[hpt!]
    \centering
    \includegraphics[width=0.85\linewidth]{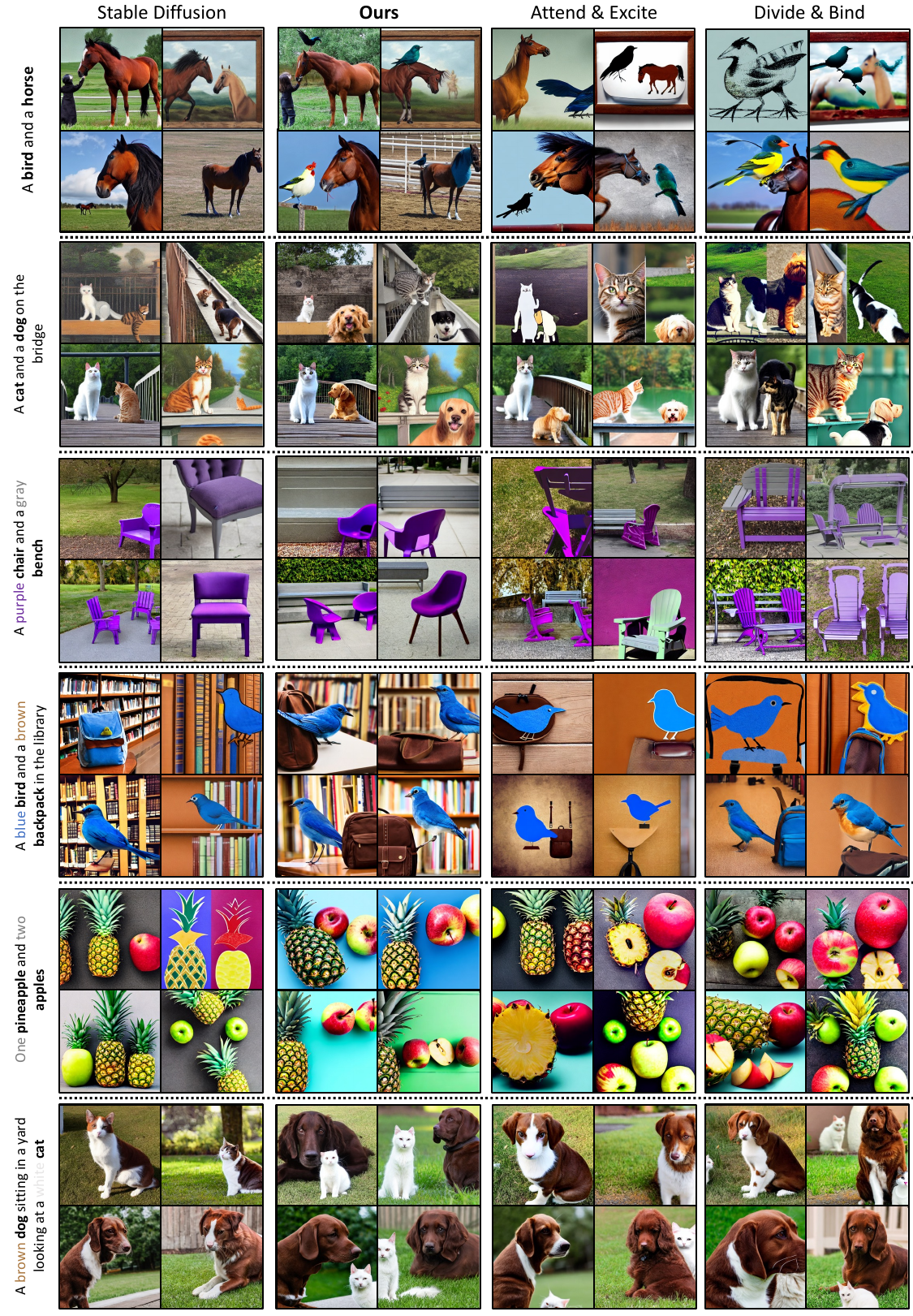}
    \caption{\textbf{Qualitative comparison of CONFORM on Stable Diffusion with other state-of-the-art methods}. Our method generates more faithful images for the input text prompt on both simple and complex prompts.}
    \label{fig:main_comparison}
    \vspace{-1em}
\end{figure*}
\subsection{CONFORM}
\label{sec:our_method}

In our approach, we utilize attention maps of object and attribute tokens as features. For a given prompt, such as `a \textcolor{BrickRed}{red \textbf{backpack}} and a \textcolor{ForestGreen}{green \textbf{suitcase}}' we group the objects and their corresponding attributes. For instance, the attention maps for \textcolor{BrickRed}{red} and \textcolor{BrickRed}{\textbf{backpack}} are grouped together, while \textcolor{ForestGreen}{green} and \textcolor{ForestGreen}{\textbf{suitcase}} are put into another group. Consequently, pairs (\textcolor{BrickRed}{red}, \textcolor{BrickRed}{\textbf{backpack}}) and (\textcolor{ForestGreen}{green}, \textcolor{ForestGreen}{\textbf{suitcase}}) are treated as positive, while pairs (\textcolor{BrickRed}{red}, \textcolor{ForestGreen}{green}), (\textcolor{BrickRed}{red}, \textcolor{ForestGreen}{\textbf{suitcase}}), (\textcolor{BrickRed}{\textbf{backpack}}, \textcolor{ForestGreen}{green}), and (\textcolor{BrickRed}{\textbf{backpack}}, \textcolor{ForestGreen}{\textbf{suitcase}}) form negative pairs. Moreover, to maintain the consistency of attention maps through successive steps in the backward diffusion process (See Fig.~\ref{fig:timestep-failure}), we also incorporate the attention maps from the timestep $t+1$ into the loss calculation, effectively doubling the token count used to calculate the loss function, creating pairs based on attention maps from the same timesteps, as well as cross-timesteps. This entails, for the color `\textcolor{BrickRed}{red}', pairs  (\textcolor{BrickRed}{red$ _t$}, \textcolor{BrickRed}{red$_{t+1}$}), and (\textcolor{BrickRed}{red$_t$}, \textcolor{BrickRed}{\textbf{backpack}$_{t+1}$}) serving as positive pairs, in addition to those formed from attention maps within the same timestep. Likewise, for the color `\textcolor{BrickRed}{red}', we introduce negative pairs like (\textcolor{BrickRed}{red$_t$}, \textcolor{ForestGreen}{green$_{t+1}$}), and (\textcolor{BrickRed}{red$_t$}, \textcolor{ForestGreen}{\textbf{suitcase}$ _{t+1}$}) to the loss calculation. For the contrastive objective, we employ InfoNCE loss, known for its fast convergence compared to previous methods. The InfoNCE loss operates on pairs of cross-attention maps, involving both object and attribute tokens from timestep $t$ and $t+1$. The loss function can be expressed for a given attention map $A^j$ as follows for a single positive pair:
\begin{equation}
    \mathcal{L} = -\log \frac{\exp(\mathrm{sim}(A^j, A^{j^+})/\tau)}
    {\sum_{n \in \{j^+, j^-_1, \cdots j^-_N\}} \exp(\mathrm{sim}(A^j, A^n)/\tau)}
    \label{eq:contrastive_loss}
\end{equation}
where $\mathrm{sim}$ function represents cosine similarity:
\begin{equation}
    \mathrm{sim}(u, v) = \frac{u^T \cdot v}{\Vert u \Vert \Vert v \Vert}
\end{equation}
In this equation, $\tau$ is the temperature parameter, and the summation in the denominator contains one positive pair and all negative pairs for $A^j$. We compute the average InfoNCE loss across all positive pairs.


\noindent \textbf{Optimization.} In our approach, the loss function consists of a single term, detailed in Section \ref{sec:our_method}. We then direct the latent representation in the desired direction as measured by the loss function. The latent representation is updated at each step as follows:
\begin{equation}
z^\prime_t = z_t - \alpha_t \nabla_{z_t} \mathcal{L}
\label{eq:optimization}
\end{equation}

 Please see the detailed algorithm in \textit{Supplementary Material}. %

\section{Experiments}
\label{sec:exp}

\begin{figure*}[t!]
    \centering
    \includegraphics[width=0.89\linewidth]{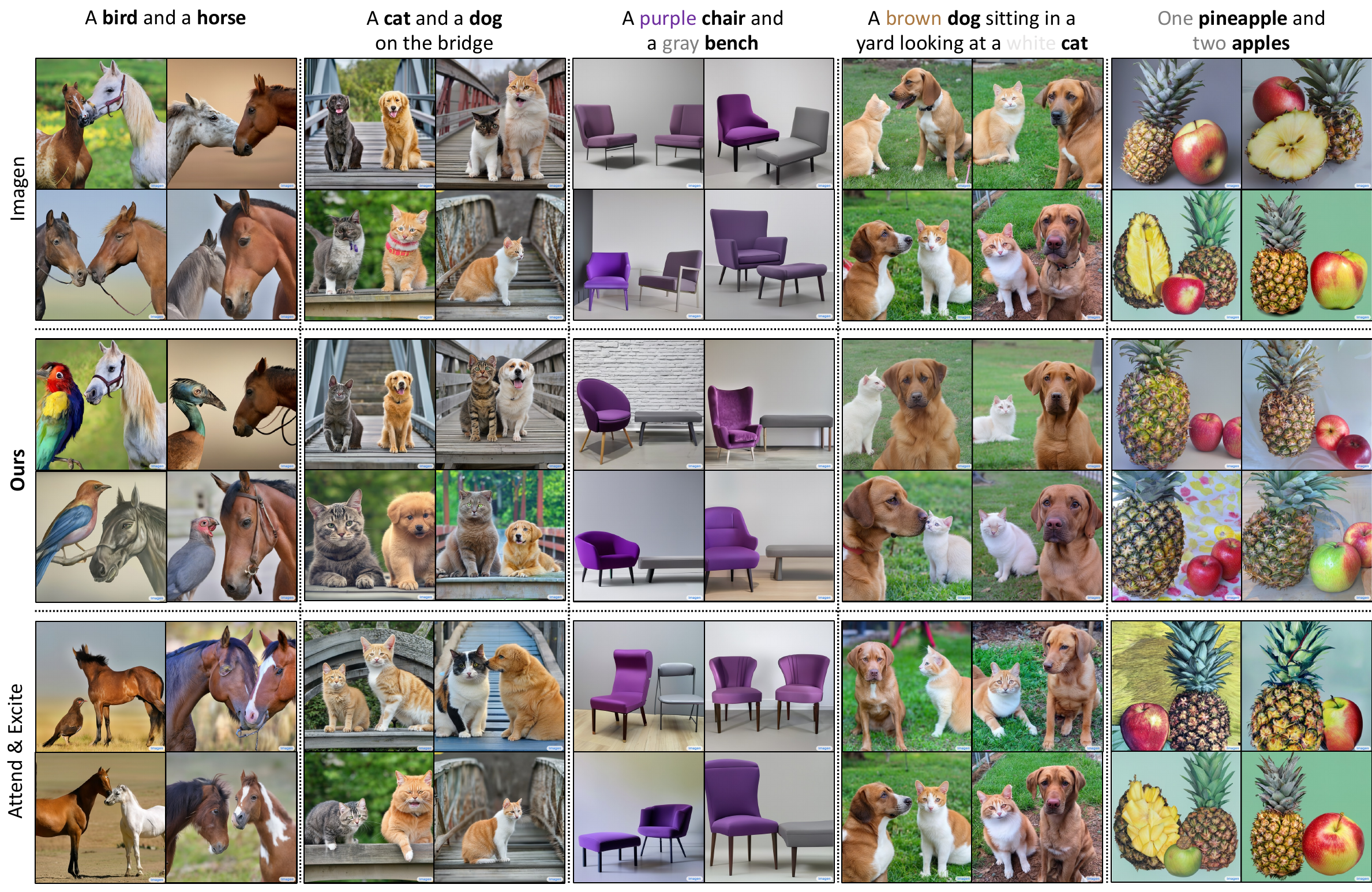}
    \caption{\textbf{Qualitative comparison of CONFORM on Imagen.} Our approach consistently produces images that more accurately reflect the input text prompts, effectively handling both simple and complex scenarios in the Imagen model. 
    }
    \label{fig:main_comparison_imagen}
    \vspace{-1em}
\end{figure*}

\paragraph{Experimental setup.} 

Due to the absence of standardized benchmarks for the evaluation of text-to-image generation models, we adopt a comprehensive evaluation strategy that combines commonly used prompts for qualitative analysis and protocols established in prior works \cite{chefer2023attend, li2023divide} for quantitative assessment. The benchmark protocol we follow comprises structured prompts `a [animalA] and a [animalB]', `a [animal] and a [color][object]', `a [colorA][objectA] and a [colorB][objectB]' \cite{chefer2023attend}, and multi-instance prompts from \cite{li2023divide}. Details of the benchmark sets and the number of prompts for each benchmark set are detailed in \textit{Supplementary Material}. For each prompt, we use 64 different seeds per prompt, utilizing 50 iterations. Using Stable Diffusion \cite{stable-diffusion} v1.5, the process takes approximately 20 seconds on an NVIDIA L4 GPU. The scale factor $\alpha$ is set to 20 (Eq.~\ref{eq:optimization}), and the temperature $\tau$ to 0.5 (Eq.~\ref{eq:contrastive_loss}). To enhance the effectiveness of our updates, we perform optimization multiple times before initiating a backward step at iterations $i \in \{0,10, 20\}$. After $i=25$, we also stop any further optimization to prevent unwanted artifacts in the output. Details for the ablation study to determine these parameters are detailed in \textit{Supplementary Material}.

\vspace{-12pt}

\paragraph{Baselines.}

We compare our results with several state-of-the-art methods, including Attend \& Excite (A\&E) \cite{chefer2023attend}, Divide \& Bind (D\&B) \cite{li2023divide}, ComposableDiffusion (ComposableD.) \cite{liu2022compositional}, and StructureDiffusion (StructureD.) \cite{feng2023structureDiffusion}. Note that while A-Star \cite{agarwal2023star} is one of our competitors, we are not able to include a comparison since their code is not available.

\begin{figure*}
    \centering
    \includegraphics[width=0.9\textwidth]{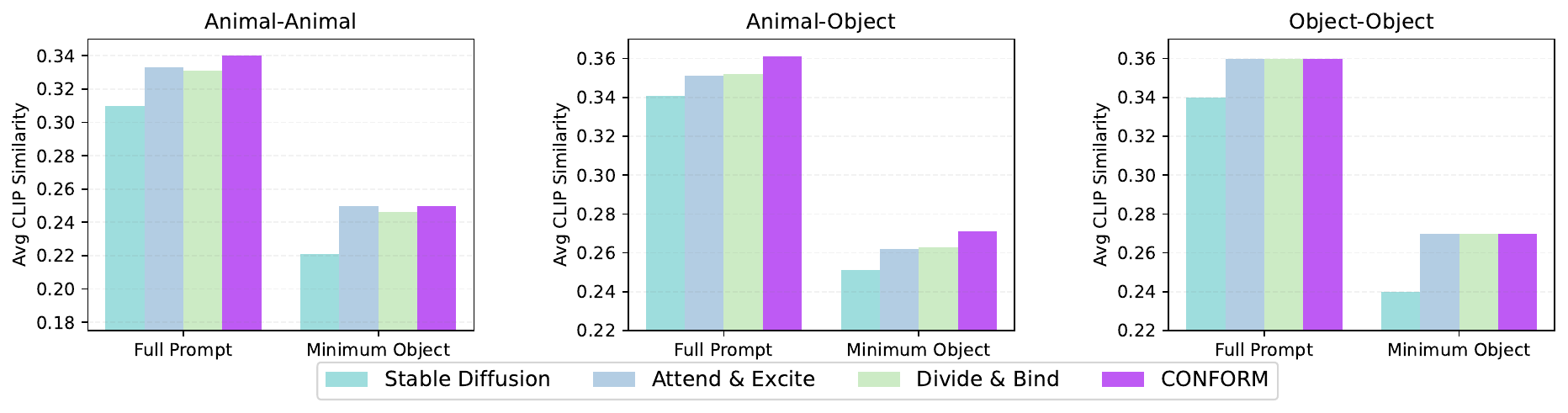}
    \vspace{-0.4cm}
    \caption{\textbf{CLIP similarity scores.} Average CLIP image-text similarities between the text prompts and the images generated by each Stable Diffusion-based method. }
    \label{fig:clip_image_similarities}
\end{figure*}
    \vspace{-0.2cm}
    
\subsection{Qualitative experiments}
\paragraph{Stable Diffusion.} Figure \ref{fig:main_comparison} presents a side-by-side comparison between CONFORM and other state-of-the-art methods using the Stable Diffusion model. Each method is evaluated using identical input seeds for consistency. CONFORM successfully addresses the issue of missing objects, as demonstrated with the \textit{`A bird and a horse'} text prompt. In scenarios where the Stable Diffusion (SD) model misses the `bird' in the image, the CONFORM method successfully integrates it, maintaining the image's original semantic integrity. Conversely, competing methods either fail to add the missing object or produce an image significantly different from the original semantic.  Our method successfully incorporates missing objects into images featuring scenes, such as the prompt \textit{`A cat and a dog on the bridge'}. Our approach effectively inserts the absent object, like a \textit{dog}, into the image. In cases where the Stable Diffusion (SD) model outputs an image with \textit{two cats}, our method can transform one of the \textit{cats} into a \textit{dog}, while preserving the original semantics of the image. In contrast, other methods either fail to respect critical scene components (\eg{,}, the \textit{bridge}) or struggle to generate the correct object. Additionally, our method can handle text prompts where objects are described with specific colors, like in \textit{`A purple chair and a gray bench'}. In such cases, the Stable Diffusion (SD) model often struggles, either failing to generate both objects simultaneously (for instance, omitting the \textit{bench}) or incorrectly assigning colors to objects. Conversely, our technique consistently produces images with both the \textit{chair} and \textit{bench}, accurately applying the designated colors (\eg{,}, \textit{purple} for the \textit{chair} and \textit{gray} for the \textit{bench}). In contrast, other methods tend to merge the colors, resulting in a bench colored in both purple and gray, or they fail to generate the \textit{bench} altogether. Our method handles attribute binding in more complex prompts involving scenes such as \textit{`A blue bird and a brown backpack in the library'} or \textit{`A brown dog sitting in a yard looking at a white cat'}. Unlike Stable Diffusion (SD) and other methods, which often struggle to produce the objects or accurately color them, our method consistently generates both the correct colors and objects. Lastly, our method manages scenarios with specific item quantities. For instance, in the text prompt \textit{`One pineapple and two apples'}, our method accurately produces the correct number of items, whereas Stable Diffusion (SD) and other methods frequently generate an excessive amount of apples.

\vspace{-12pt}

\paragraph{Imagen.} Additionally, the effectiveness of our method with Imagen is demonstrated in Fig.~\ref{fig:main_comparison_imagen}. Our primary comparison is against Attend \& Excite, adapted to maximize the presence of multiple tokens constituting the target word. Given Imagen's use of the T5 model and its tendency to split words into tokens (\eg{,}, \textit{zebra} becomes \textit{ze, bra}), it is not clear how Divide \& Bind approach, specifically the attribute binding regularization, can be applied to Imagen. Notably, CONFORM naturally handles such situations by treating the attention maps of tokens like \textit{ze} and \textit{bra} as positive pairs.  Our findings reveal that our method successfully addresses the issue of missing objects, as seen in the `\textit{A bird and a horse}' prompt. Where Imagen originally failed to generate a bird and produced two horses instead, our method effectively substitutes a horse for a bird while maintaining the original semantics of the image. In contrast, Attend \& Excite often either fails to generate the image or significantly alters the scene. Similarly, our method successfully handles prompts like \textit{`A cat and a dog on the bridge'}, where Imagen or Attend \& Excite typically result in images of two cats; our method replaces one of the cats with a dog. For text prompts involving specific colors, like \textit{`A purple chair and a gray bench'} and \textit{`A brown dog sitting in a yard looking at a white cat'}, our method accurately assigns the colors to the appropriate objects. In contrast, Imagen and Attend \& Excite struggle with these tasks, often failing to produce a bench or incorrectly coloring the objects. Lastly, our method successfully generates the accurate number of objects for \textit{`One pineapple and two apples'} prompt, while other methods fail to generate the correct number of apples.

\subsection{Quantitative experiments}

\begin{table}
    \small
    \centering
    \setlength{\tabcolsep}{2pt}
    \caption{Average CLIP text-text similarities between the text prompts and captions generated by BLIP for Stable Diffusion-based methods. \\[-0.65cm]} 
    \begin{tabular}{l c c c} 
        \toprule
        Method & Animal-Animal & Animal-Object &  Object-Object  \\
        \midrule
        SD             & 0.76  & 0.78   & 0.77   \\
        ComposableD.      & 0.69  & 0.77   & 0.76   \\
        StructureD. & 0.76  & 0.78   & 0.76   \\
        A\&E             & 0.80  & 0.83   & 0.81  \\
        D\&B                   & 0.81  & 0.83    & 0.81   \\
        \textbf{CONFORM}                        & \textbf{0.82} & \textbf{0.85} & \textbf{0.82}  \\
        \bottomrule \\[-0.8cm]
    \end{tabular}
    \label{tb:blip_captioning_similarity}
\end{table}

\begin{table}[t]
    \small
    \centering
    \setlength{\tabcolsep}{2pt}
    \caption{Average TIFA scores for SD and Imagen.
    \\[-0.25cm]} 
    \begin{tabular}{l c c c c} 
        \toprule
        Method &  Animal-Animal  &  Animal-Obj  & Obj-Obj & Multi-Obj \\
        \midrule
        SD             & 0.68  & 0.80   & 0.65  & 0.59 \\
        A\&E             & 0.92  & 0.91   & 0.82  & 0.72\\
        D\&B                  & 0.93  & 0.91    & 0.83   & 0.73\\
        \textbf{CONFORM}                        & \textbf{0.95} & \textbf{0.94} & \textbf{0.88} & \textbf{0.74}  \\
        \toprule
        Imagen                       & \textbf{0.84}  & 0.93   & 0.88  & 0.73 \\
        A\&E             & \textbf{0.84}  & 0.93   & 0.88  & 0.73 \\
        \textbf{CONFORM}             & \textbf{0.84} & \textbf{0.94} & \textbf{0.91} & \textbf{0.76}  \\
        \bottomrule \\[-0.45cm]
    \end{tabular}
    \label{tb:tifa_similarity}
\end{table}
 
To quantitatively assess the efficacy of our approach, we employ multiple metrics, including image-text similarity, text-text similarity, and the recently introduced TIFA score \cite{hu2023tifa}. We assess image-text similarity using the CLIP similarity metric, comparing the generated image with the input prompt. We calculate both the full-prompt similarity (CLIP-full), representing the likeness between the entire prompt and the generated image, and the minimum object similarity (CLIP-min), which is the minimum of the similarities between the generated image and each of the two subject prompts. It is noteworthy that while our model achieved comparable or higher results compared to the reference methods, these metrics should be interpreted with caution, as the models used for comparison are already conditioned on CLIP embeddings. In direct comparison with Stable Diffusion, our method outperformed in both CLIP-full and CLIP-min similarity scores across most of the benchmark sets while performing similarly at others (see Fig. \ref{fig:clip_image_similarities}).

For text-text similarity, we leverage BLIP \cite{li2022blip} to generate captions for the generated image. Then, we evaluate the similarity between the input prompt and these captions. This assessment aims to capture subjects and attributes present in the original prompt, providing insights into the coherence and relevance of the textual descriptions. In comparative analysis with Stable Diffusion and other competitors, our method consistently demonstrated superior performance in text-text similarity scores across all benchmark sets (see Tab. \ref{tb:blip_captioning_similarity}). 

The TIFA score \cite{hu2023tifa} provides an evaluation by assessing the faithfulness of the generated image to the input prompt. To compute the TIFA score, we automatically generate a set of question-answer pairs using the GPT-3.5 \cite{brown2020language} language model. Image faithfulness is then determined by evaluating the proportion of correct answers using the visual question answering model UnifiedQA-v2 \cite{khashabi2022unifiedqa}. This metric offers a comprehensive evaluation by considering both textual and visual aspects of the generated content. Our method consistently outperforms across all benchmark sets in the TIFA metric in SD. In addition, CONFORM outperforms Attend \& Excite and Imagen in most of the benchmarks while performing similarly at the `Animal-animal' benchmark  (see Tab. \ref{tb:tifa_similarity}).

\vspace{-12pt}

\paragraph{User study.} To evaluate the fidelity of images generated by our model, we conducted a user study involving 25 participants. We selected 10 random prompts and generated four images for each using different seeds. This process was repeated separately for both Stable Diffusion-based and Imagen-based models. Participants were asked to choose the image reflecting the text prompt best for each combination of prompt and seed. Results, detailed in Tab. \ref{tb:user_study}, overwhelmingly favored CONFORM. For Stable Diffusion, CONFORM led in all categories, achieving 72\% to 94\% of the votes across different benchmark sets. For Imagen study, it similarly dominated, receiving 96\% to 98\% of the votes. These results highlight CONFORM's effectiveness in closely aligning generated images with the text prompts.

\begin{figure}[t]
    \centering
    \includegraphics[width=1\linewidth]{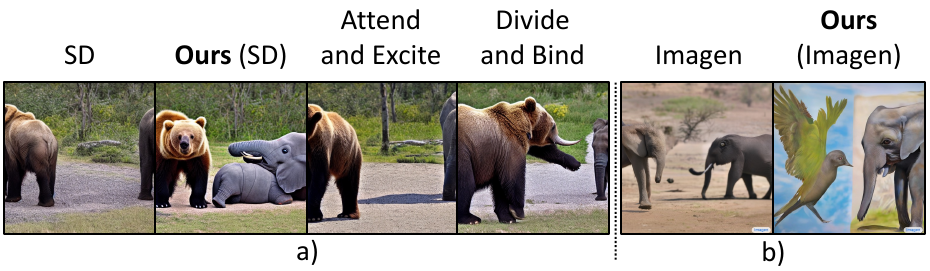}
    \caption{\textbf{Limitations.} (a) Shows challenges when key objects are missing from SD image for the `A bear and an elephant' prompt. (b) Displays object separation in Imagen, despite enhanced prompt accuracy for the `A bird and an elephant' prompt.}
    \label{fig:limitations}
\end{figure}

\begin{table}
    \small
    \centering
    \setlength{\tabcolsep}{2pt}
    \caption{User study with $25$ respondents for SD  and Imagen. \\[-0.25cm]} 
    \begin{tabular}{l c c c c} 
        \toprule
        Method &  Animal-Animal  &  Animal-Obj  & Obj-Obj & Multi-Obj \\
        \midrule
        SD       & 5\%           & 3\%           & 0\%           & 5\% \\
        A\&E     & 14.75\%       & 2\%           & 7.5\%           & 2\% \\
        D\&B         & 8.25\%        & 1\%           & 4.5\%           & 2\% \\
        \textbf{CONFORM}                & \textbf{72\%} & \textbf{94\%}  & \textbf{88\%}  &\textbf{91\%} \\
        \toprule
        Imagen        & 3.25\%        & 4\%           & 2\%           & 3.5\% \\
        \textbf{CONFORM}                & \textbf{96.75\%} & \textbf{96\%}  & \textbf{98\%}  &\textbf{96.5\%} \\
        \bottomrule
    \end{tabular}
    \vspace{-0.2cm}
    \label{tb:user_study}
\end{table}

\section{Limitations}

A  limitation of our method based on SD: when the initial map significantly excludes objects, ours may struggle to generate successful images, although it is still able to place the desired objects into the generated image (Fig.~\ref{fig:limitations} (a)). This issue does not apply to our method; others also encounter difficulties when starting with challenging attention maps. In Imagen, our refinement process might sometimes lead to the separation of objects, yet it still enhances the accuracy of the text prompt in the final image (Fig.~\ref{fig:limitations} (b)).

\section{Conclusion}
\label{sec:conclusion}

In our study, we introduced a novel framework centered on a contrastive objective, designed to enhance the fidelity of text-to-image diffusion models. Our approach is model-agnostic and applied to popular text-to-image generators like Stable Diffusion and Imagen. Through comprehensive experiments on multiple benchmark datasets, we assessed our method using text-image similarity, text-text similarity, and TIFA scores, comparing it with several leading techniques. Our findings reveal that our method consistently produces images that are more faithful to the original text prompts than the baseline methods in both Stable Diffusion and Imagen models.

\newpage
\clearpage
{
    \small
    \bibliographystyle{ieeenat_fullname}
    \bibliography{main}
}

\clearpage
\maketitlesupplementary

\section{Benchmark Sets}

In our qualitative analysis approach, we adopt benchmark sets from previous studies \cite{chefer2023attend, li2023divide}. Our benchmark protocol includes structured prompts such as `an [animalA] and an [animalB]', `an [animal] with a [color] [object]', and `a [colorA] [objectA] and a [colorB] [objectB]', along with multi-instance prompts. Table \ref{tab:benchmark_sets} provides details on the benchmark sets and the number of prompts for each set. We test each prompt with 64 unique seeds, conducting 50 iterations per seed. The testing is performed using Stable Diffusion v1.5 and typically takes about 20 seconds per prompt on an NVIDIA L4 GPU.
 
\begin{table}[h!]
    \centering
    \caption{Description of benchmark sets and number of prompts used for qualitative evaluation.} %
    \begin{tabular}{lcr}
        \toprule
        Benchmark Set & Template & \#\\
        \midrule
        \multirow{ 2}{*}{Animal-Animal} & a [animalA] and a [animalB] & \multirow{ 2}{*}{66}\\ 
        & \textit{`a \underline{\textbf{horse}} and a \underline{\textbf{bird}}'} \\ \\
        \multirow{3}{*}{Animal-Object} & a [animal] and    & \multirow{ 3}{*}{144 }\\
                                        & a [color][object] & \\
         & \textit{`a \underline{\textbf{frog}} and a \underline{\textbf{purple balloon}}'} \\ \\
        \multirow{3}{*}{Object-Object} & a [colorA][objectA] and  & \multirow{ 3}{*}{66 }\\
                                        & a [colorB][objectB]  &\\
         & \textit{`a \underline{\textbf{black crown}} and a \underline{\textbf{red car}}'} \\ \\
        \multirow{3}{*}{Multi-Object}  &   [numberA][animalA] and & \multirow{ 3}{*}{30 }\\
                                        & [numberB][animalB]  &\\
         & \textit{`\underline{\textbf{one zebra}} and \underline{\textbf{two birds}}'} \\
        \bottomrule \\[-0.8cm]
    \end{tabular}
    \label{tab:benchmark_sets}
\end{table}

\section{Algorithm}
\label{sec:algorithm}

\algnewcommand{\algorithmicgoto}{\textbf{Go to}}%
\algnewcommand{\Goto}[1]{\algorithmicgoto~\ref{#1}}%

\begin{algorithm}[t!]
\caption{A Single Denoising Step using CONFORM.}

\begin{flushleft}
\textbf{Input:} A text prompt $\mathcal{P}$, previous attention maps $A_{t+1}$, a dictionary of subject and corresponding attribute token indices $\mathcal{T}$, a timestep $t$, a set of iterations for refinement $\{t_1,\dots,t_k\}$, and a pretrained diffusion model $\epsilon_{\theta}$.\\
\textbf{Output:} A noised latent $z_{t-1}$ for the next timestep.
\end{flushleft}

\begin{algorithmic}[1]
\State $\_, A_t \gets \epsilon_{\theta}(z_t, \mathcal{P}, t)$
\State $\mathrm{L} \gets \{\}$ \Comment{\small{Pseudo-Labels}}
\State $\mathrm{F} \gets \{\}$ \Comment{\small{Features}}

\For{$s_i, \mathcal{C} \in \mathcal{T}$} \Comment{\small{Prepare features and labels}}
    \State $\mathrm{L} \gets \mathrm{L} + i$    \Comment{\small{Label attention map}}
    \State $\mathrm{F} \gets \mathrm{F} + A_t[:, :, s_i]$   \Comment{\small{Get attention map}}
    \State $\mathrm{L} \gets \mathrm{L} + i$    \Comment{\small{Label prev attention map}}
    \State $\mathrm{F} \gets \mathrm{F} + A_{t+1}[:, :, s_i]$ \Comment{\small{Get prev attention map}}
    \For{$c_j \in \mathcal{C}$} \Comment{\small{Attributes of a token get the same labels}}
        \State $\mathrm{L} \gets \mathrm{L} + i$    \Comment{\small{Label attention map}}
        \State $\mathrm{F} \gets \mathrm{F} + A_t[:, :, c_j]$   \Comment{\small{Get attention map}}
        \State $\mathrm{L} \gets \mathrm{L} + i$    \Comment{\small{Label prev attention map}}
        \State $\mathrm{F} \gets \mathrm{F} + A_{t+1}[:, :, c_j]$ \Comment{\small{Get prev attention map}}
    \EndFor
\EndFor

\State $\mathcal{L} \gets \mathcal{L}(\mathrm{F}, \mathrm{L})$
\State $z_t' \gets z_t - \alpha_t \cdot \nabla_{z_t} \mathcal{L}$

\If{$t \in \{t_1, \dots, t_k\}$} \Comment{\small{If performing iterative refinement at $t$}}
    \State $z_t \gets z_t'$
    \State \textbf{Go to} Step 1
\EndIf

\State $z_{t-1}, \_ \gets \epsilon_{\theta}(z_t', \mathcal{P}, t)$
\State \textbf{Return} $z_{t-1}$

\end{algorithmic}
\label{alg:conform}
\end{algorithm}
The pseudocode for the CONFORM algorithm is described in Algorithm \ref{alg:conform}. 

\section{Ablation Study}

In our ablation study, we conducted targeted experiments on a limited set of prompts and a smaller number of images to identify the most effective parameters and techniques for our final model, guided by CLIP similarity metrics. The most influential parameters were found to be: incorporating attention maps from previous iterations, fine-tuning the temperature parameter in the contrastive loss equation (referenced in Eq. \ref{eq:contrastive_loss}), implementing refinement steps for iterative updates of latent before advancing to subsequent iterations, and determining whether to consistently apply optimization throughout all iterations or to stop it at a certain point.

\paragraph{Using Attention Maps from Previous Iterations.} As indicated in Tab. \ref{tb:using_previous_iteration}, employing attention maps from the previous iteration positively impacts both CLIP-full and CLIP-min metrics. The benefits of this technique are also evident in Fig. \ref{fig:timestep-failure} (see main paper). While the diffusion process naturally leads to the scattering of attention, our method assists in maintaining focused attention throughout the iterations, countering the scattering effect.

\begin{table}[h]
    \small
    \centering
    \setlength{\tabcolsep}{2pt}
    \caption{Average CLIP image-text similarities for ablation study on the effect of using previous iteration attention maps. \\[-0.65cm]} 
    \begin{tabular}{l c c c c} 
        \toprule
        Method  & CLIP-full &  CLIP-min \\
        \midrule
        Stable Diffusion             & 0.33  & 0.24    \\
        w/o Previous Iteration Attention Maps            & 0.35  & 0.25    \\
        \textbf{CONFORM}             & \textbf{0.36} & \textbf{0.26} \\
        \bottomrule \\[-0.8cm]
    \end{tabular}
    \label{tb:using_previous_iteration}
\end{table}

\paragraph{Scale Factor, Temperature, Refinement Steps, Maximum Optimization Steps} The temperature parameter ($\tau$) in Eq. \ref{eq:contrastive_loss} (see main paper) is critical for achieving desirable results and requires precise calibration. We utilized a grid search ($\tau \in \{0.25, 0.5, 0.75, 1.0\}$) to identify the optimal temperature parameter and the number of refinement steps at specific iterations. Additionally, we determined the appropriate point to cease optimization and the scale factor $\alpha$, as defined in Eq. \ref{eq:optimization}. The selection was based on achieving the highest CLIP-full and CLIP-min similarity scores.

Beyond quantitative analysis, we carefully observed how each parameter influenced the final image output. Values of $\tau$ that are too high result in negligible changes, while values that are too low lead to unwanted artifacts. We determined $\tau =0.5$ as the optimal setting. Implementing refinement steps at certain iterations, specifically at $i \in {0, 10, 20}$ (similar to the approach in \cite{chefer2023attend}), allows the model to make necessary adjustments to the attention maps. However, applying optimization at every step led to unwanted artifacts in the output. Consequently, we decided to halt optimization at $i=25$ to avoid these issues. Lastly, we also used scale factor $\alpha = 20$ considering the results of the grid search.

\section{Additional Qualitative Results}

In the remainder of the Supplementary Materials, we provide additional qualitative comparisons:

\begin{itemize}
    \item Comparison with the A-Star for the images presented in \cite{agarwal2023star}.
    \item Additional qualitative comparisons for SD-based models: Stable Diffusion, Attend \& Excite, Divide \& Bind, and our method applied to the Stable Diffusion model.
    \item Additional comparisons between Imagen and our method applied to the original Imagen model.
\end{itemize}

\begin{figure*}[t!]
    \centering
    \includegraphics[width=0.89\linewidth]{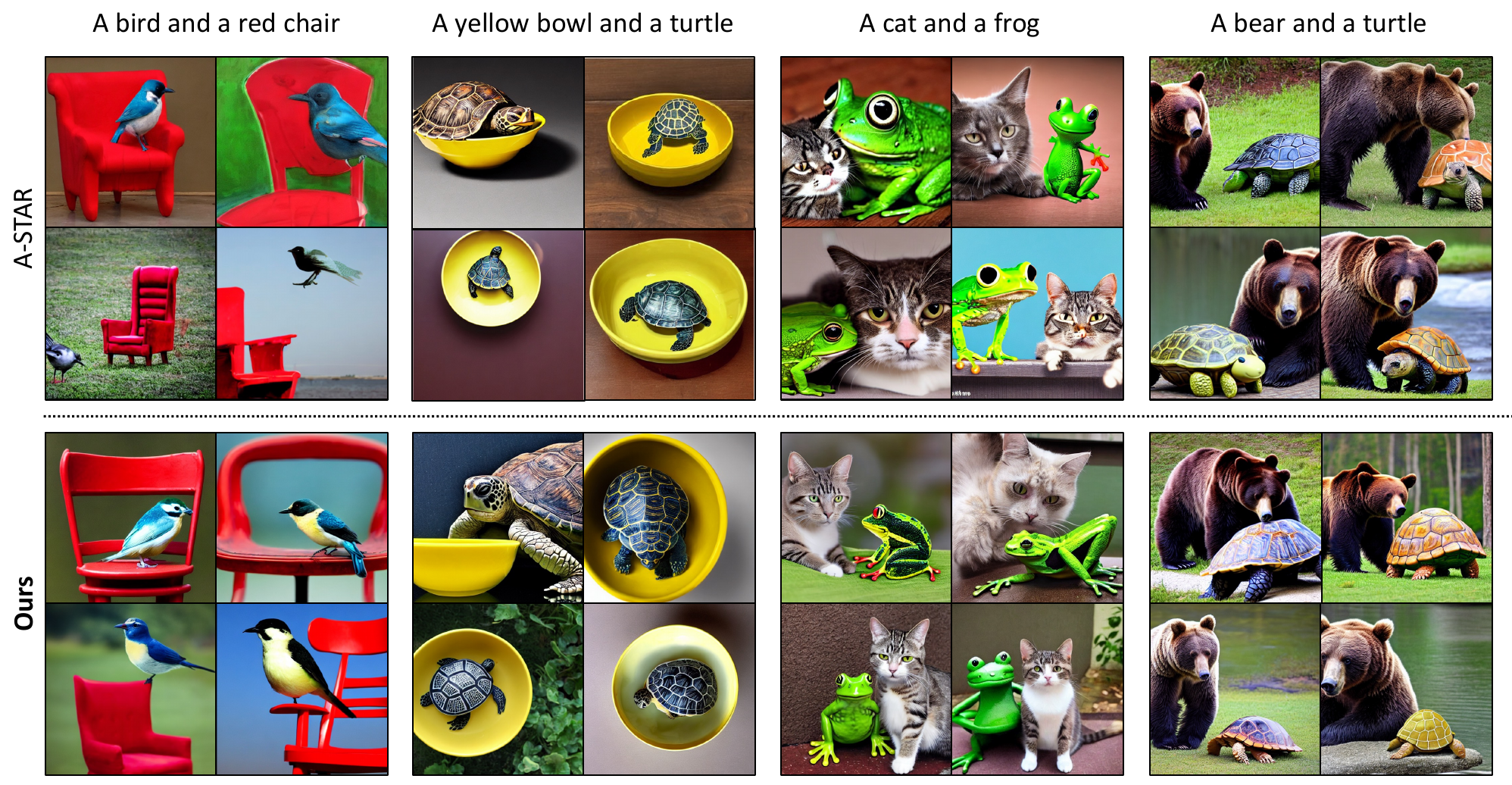}
    \caption{\textbf{Qualitative comparison of CONFORM with A-Star.} We compared the results presented in A-Star \cite{agarwal2023star} paper with our method on Stable Diffusion.}
    \label{fig:astar_comparison}
    \vspace{-1em}
\end{figure*}

\begin{figure*}[t!]
    \centering
    \includegraphics[width=0.89\linewidth]{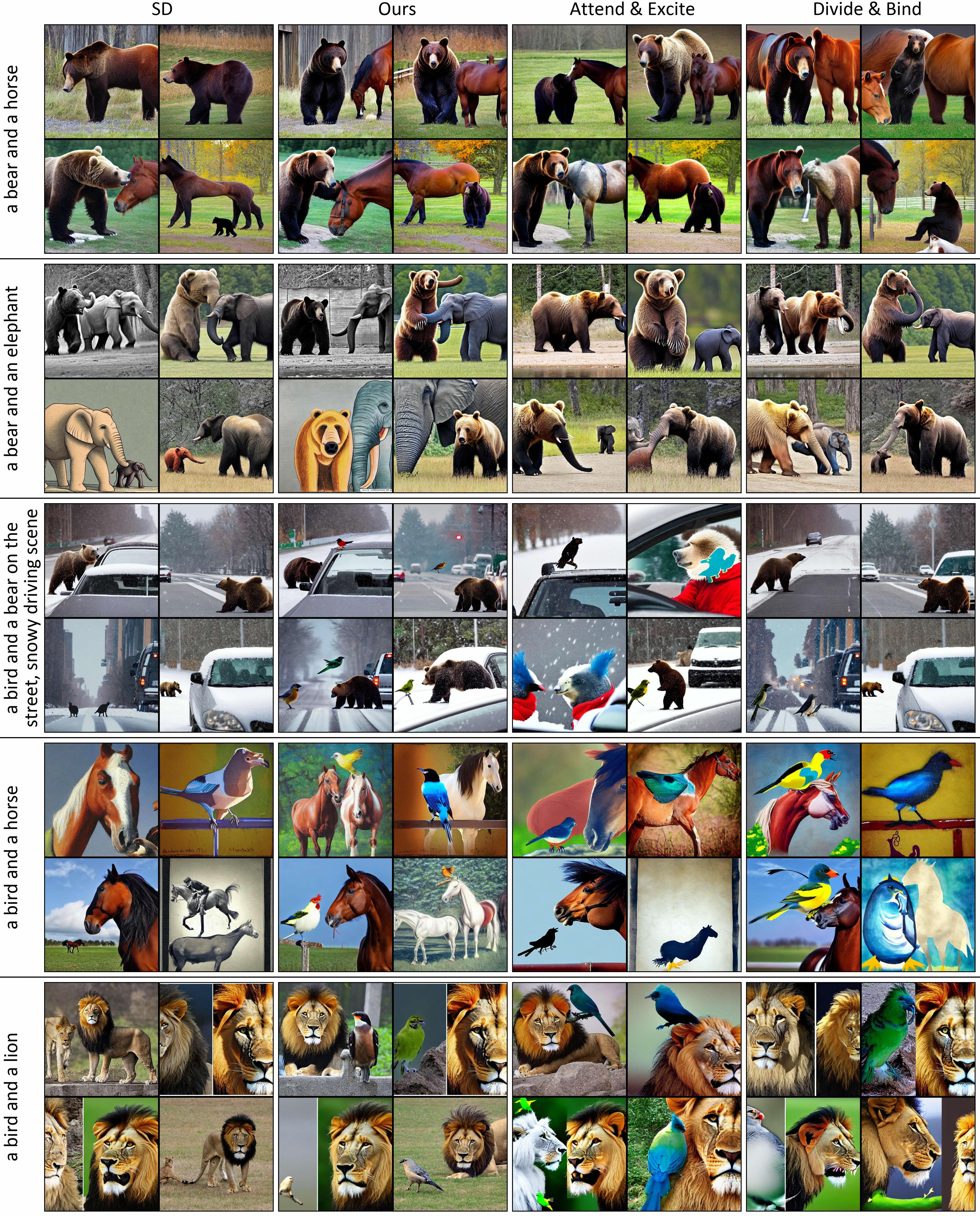}
    \caption{\textbf{Qualitative comparison of CONFORM on SD.} Our approach consistently produces images that more accurately reflect the input text prompts, effectively handling both simple and complex scenarios in the SD model. 
    }
    \label{fig:supplementary_sd_page_0}
    \vspace{-1em}
\end{figure*}

\begin{figure*}[t!]
    \centering
    \includegraphics[width=0.89\linewidth]{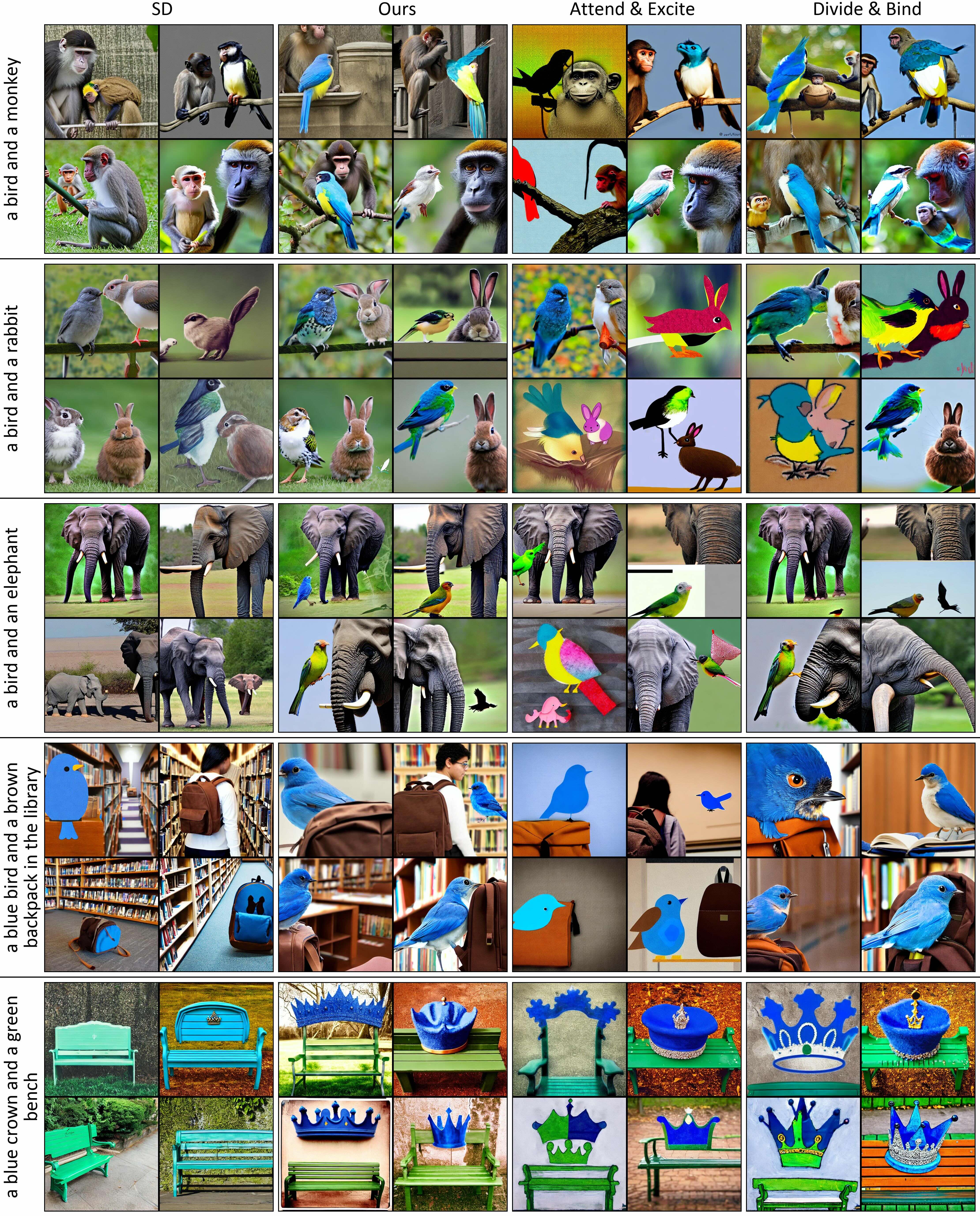}
    \caption{\textbf{Qualitative comparison of CONFORM on SD.} Our approach consistently produces images that more accurately reflect the input text prompts, effectively handling both simple and complex scenarios in the SD model. 
    }
    \label{fig:supplementary_sd_page_1}
    \vspace{-1em}
\end{figure*}

\begin{figure*}[t!]
    \centering
    \includegraphics[width=0.89\linewidth]{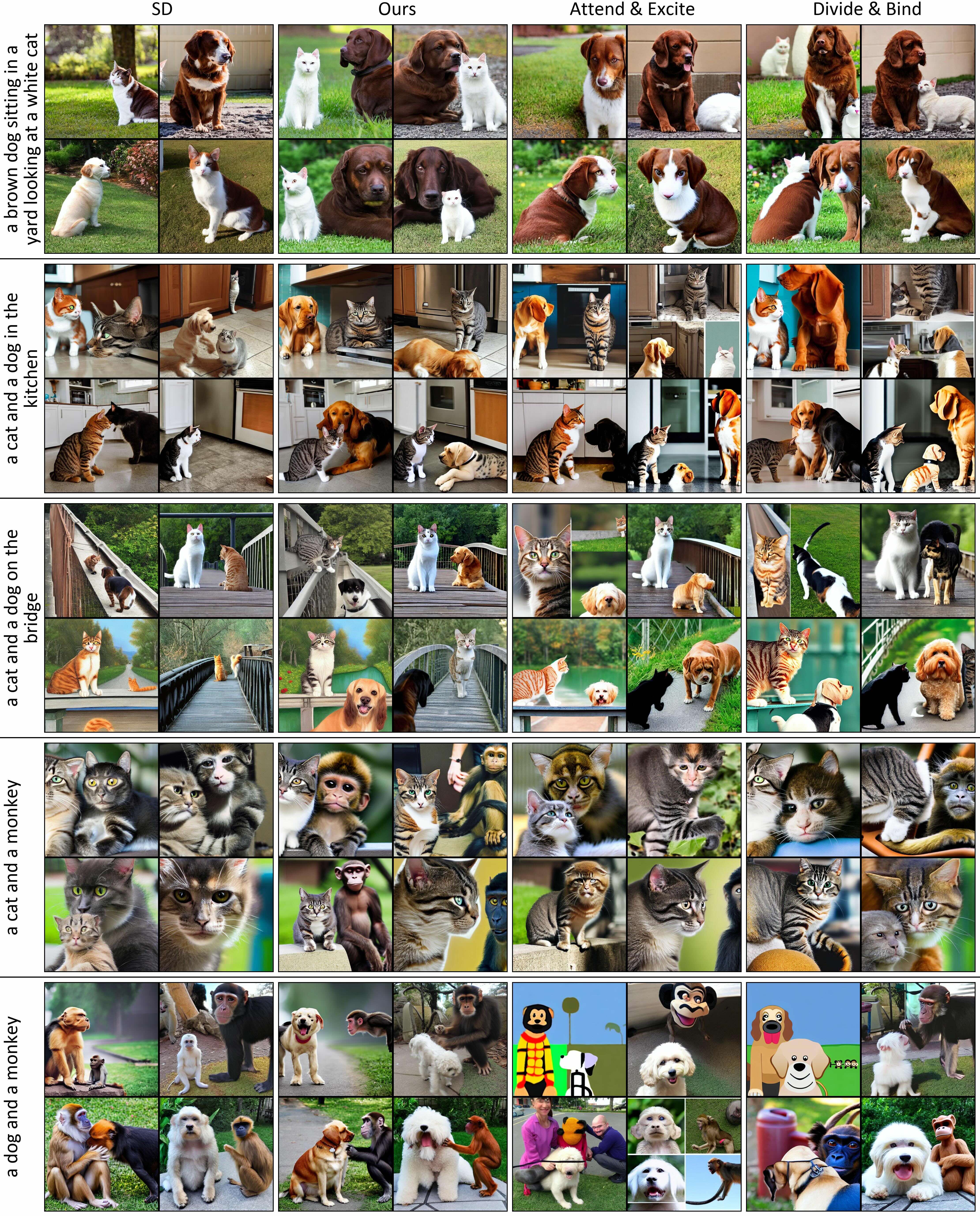}
    \caption{\textbf{Qualitative comparison of CONFORM on SD.} Our approach consistently produces images that more accurately reflect the input text prompts, effectively handling both simple and complex scenarios in the SD model. 
    }
    \label{fig:supplementary_sd_page_2}
    \vspace{-1em}
\end{figure*}

\begin{figure*}[t!]
    \centering
    \includegraphics[width=0.89\linewidth]{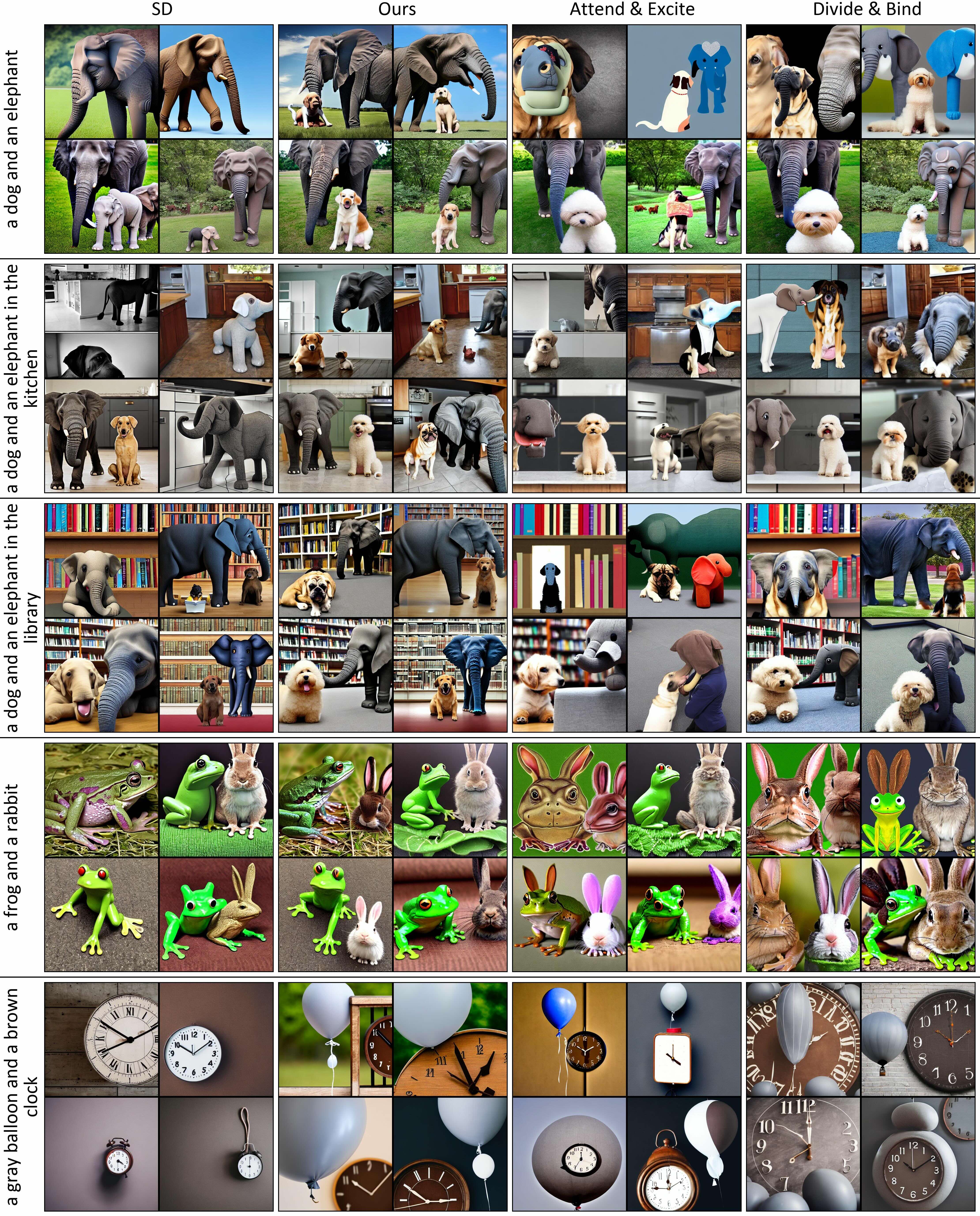}
    \caption{\textbf{Qualitative comparison of CONFORM on SD.} Our approach consistently produces images that more accurately reflect the input text prompts, effectively handling both simple and complex scenarios in the SD model. 
    }
    \label{fig:supplementary_sd_page_3}
    \vspace{-1em}
\end{figure*}

\begin{figure*}[t!]
    \centering
    \includegraphics[width=0.89\linewidth]{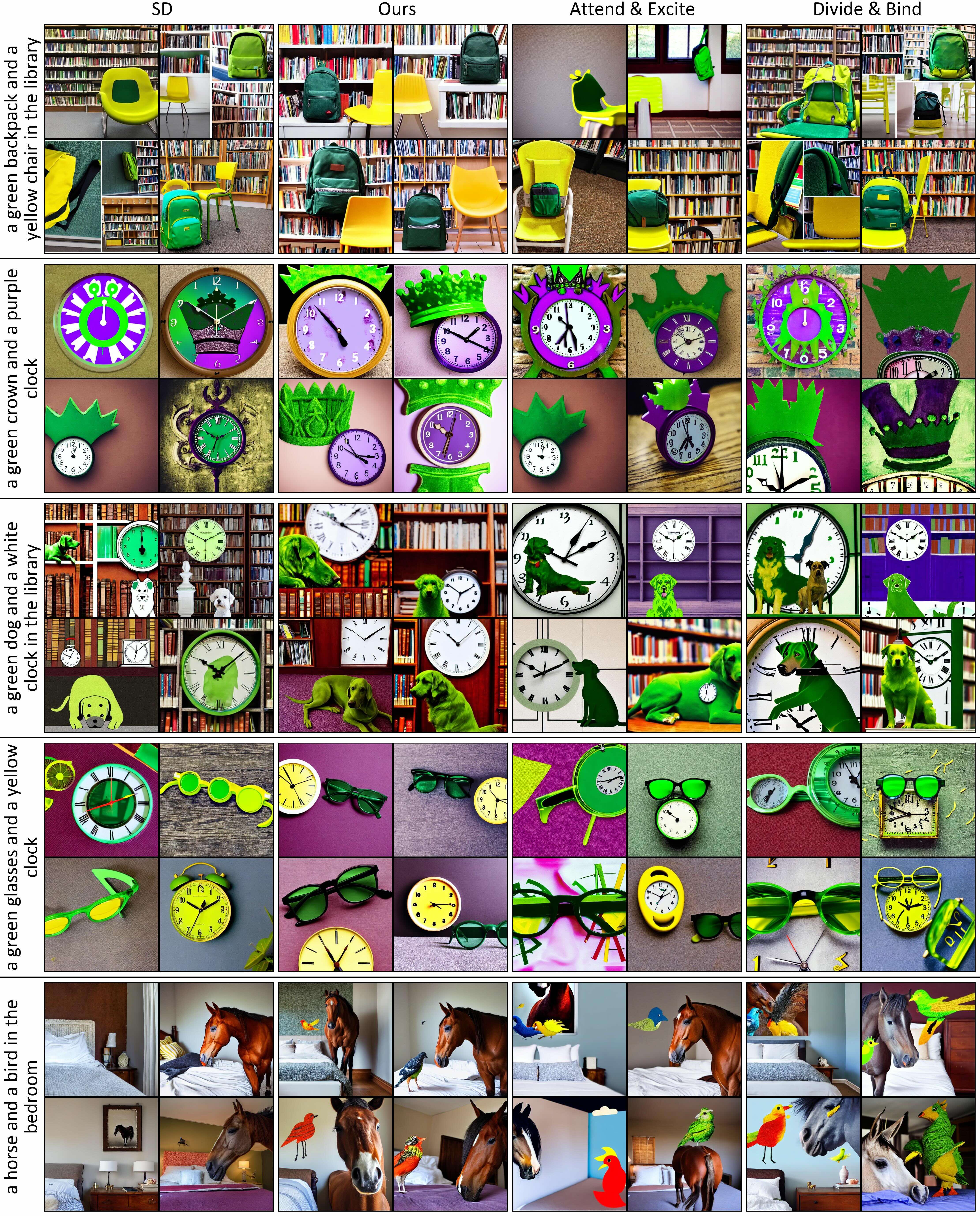}
    \caption{\textbf{Qualitative comparison of CONFORM on SD.} Our approach consistently produces images that more accurately reflect the input text prompts, effectively handling both simple and complex scenarios in the SD model. 
    }
    \label{fig:supplementary_sd_page_4}
    \vspace{-1em}
\end{figure*}

\begin{figure*}[t!]
    \centering
    \includegraphics[width=0.89\linewidth]{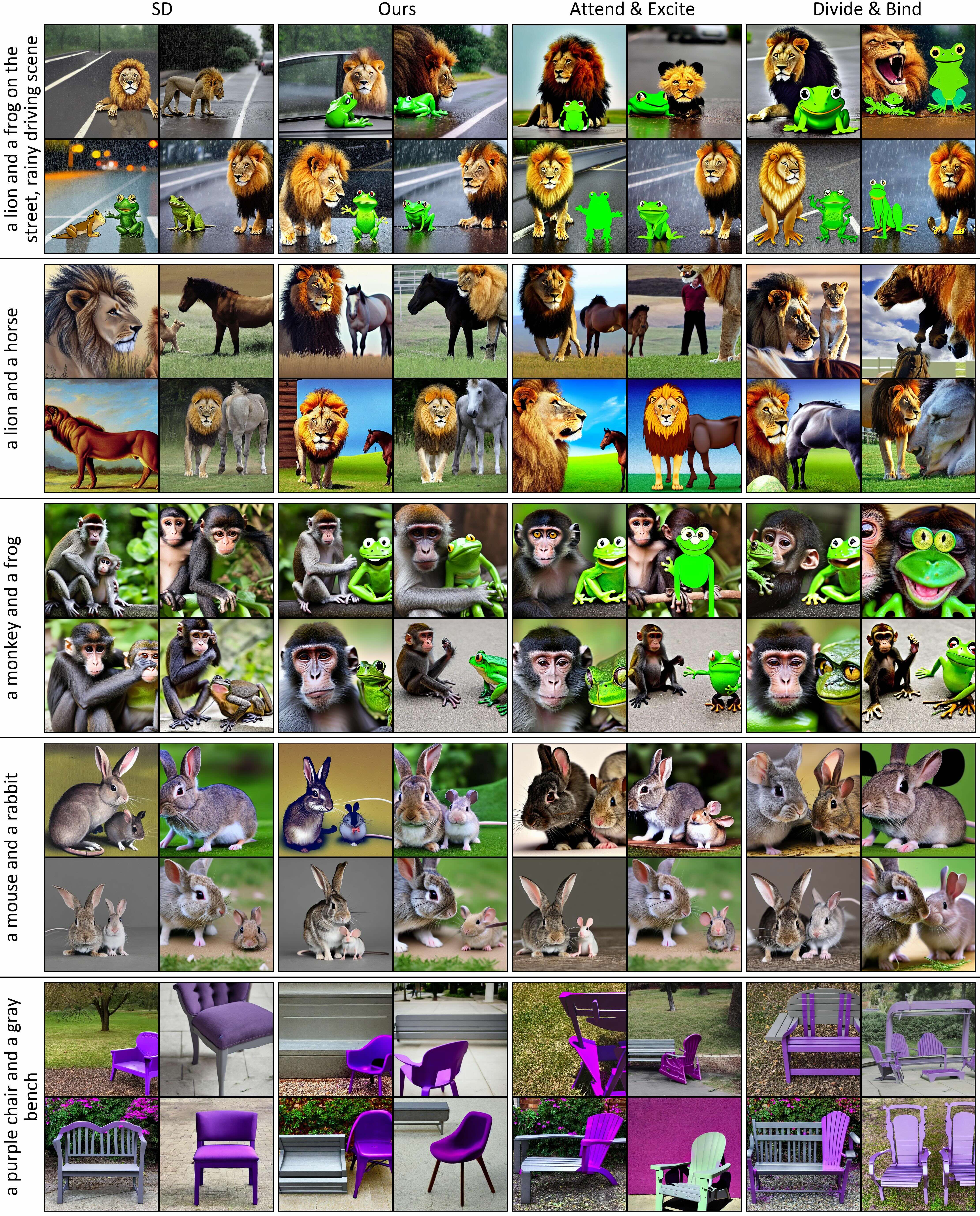}
    \caption{\textbf{Qualitative comparison of CONFORM on SD.} Our approach consistently produces images that more accurately reflect the input text prompts, effectively handling both simple and complex scenarios in the SD model. 
    }
    \label{fig:supplementary_sd_page_5}
    \vspace{-1em}
\end{figure*}

\begin{figure*}[t!]
    \centering
    \includegraphics[width=0.89\linewidth]{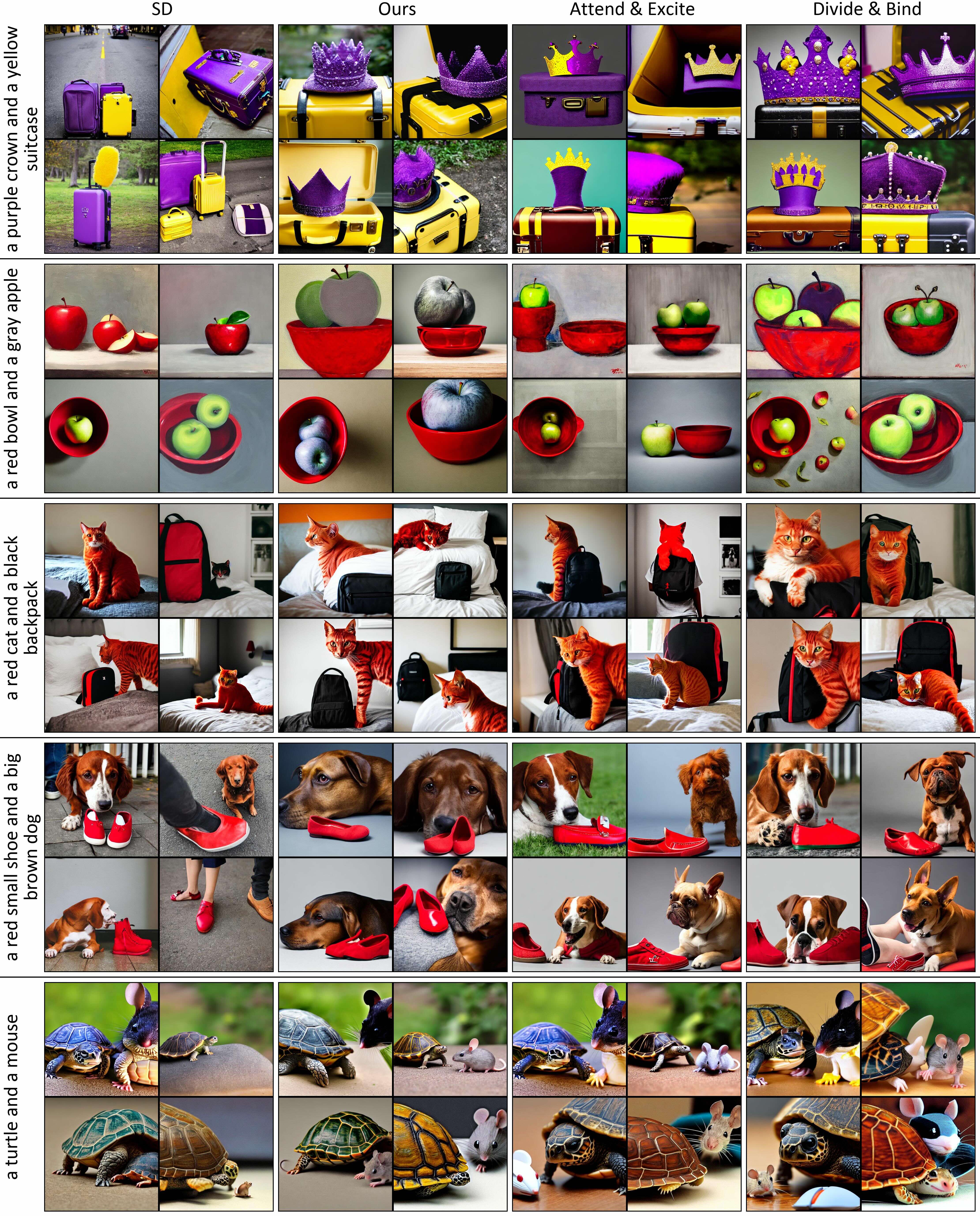}
    \caption{\textbf{Qualitative comparison of CONFORM on SD.} Our approach consistently produces images that more accurately reflect the input text prompts, effectively handling both simple and complex scenarios in the SD model. 
    }
    \label{fig:supplementary_sd_page_6}
    \vspace{-1em}
\end{figure*}

\begin{figure*}[t!]
    \centering
    \includegraphics[width=0.89\linewidth]{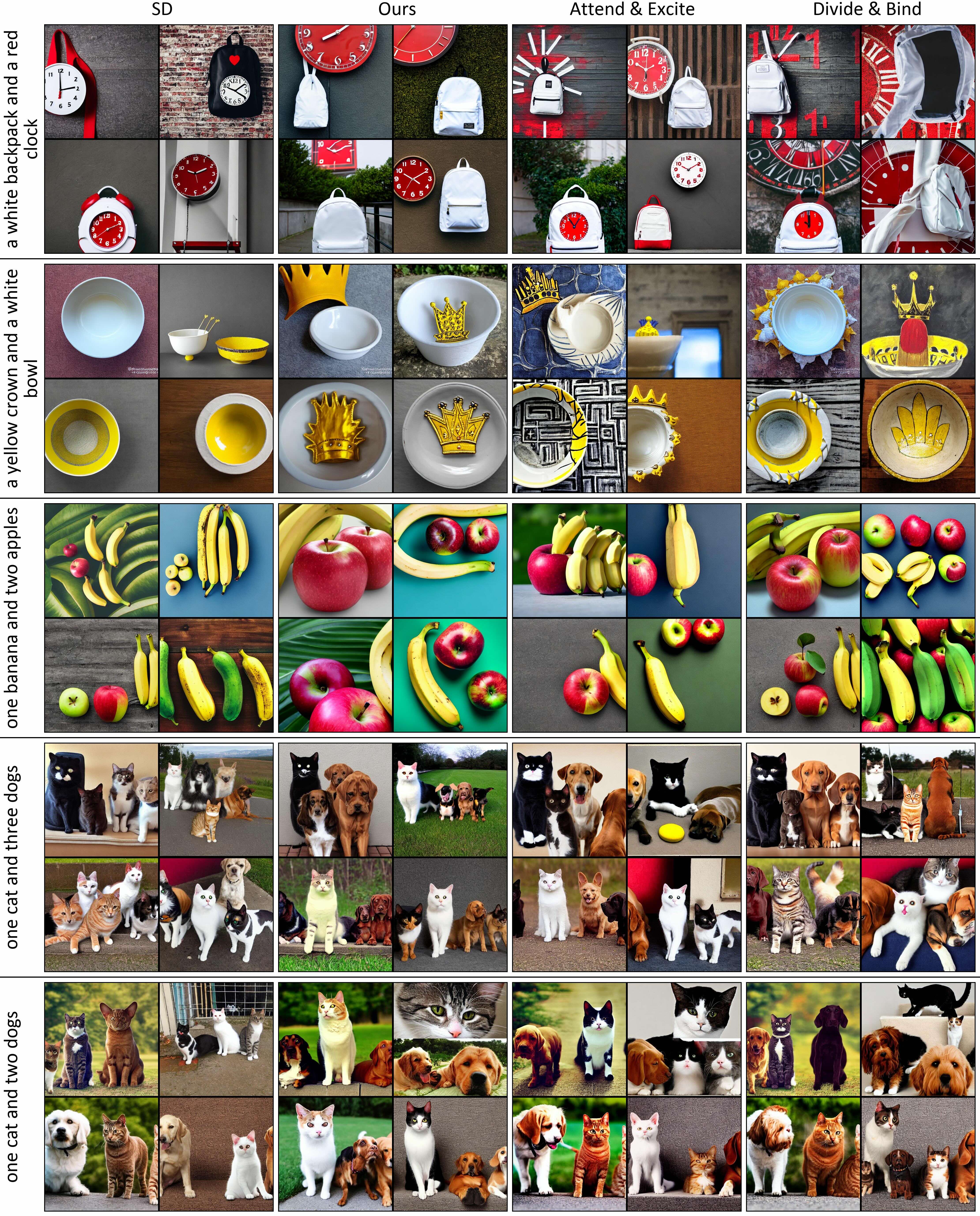}
    \caption{\textbf{Qualitative comparison of CONFORM on SD.} Our approach consistently produces images that more accurately reflect the input text prompts, effectively handling both simple and complex scenarios in the SD model. 
    }
    \label{fig:supplementary_sd_page_7}
    \vspace{-1em}
\end{figure*}

\begin{figure*}[t!]
    \centering
    \includegraphics[width=0.89\linewidth]{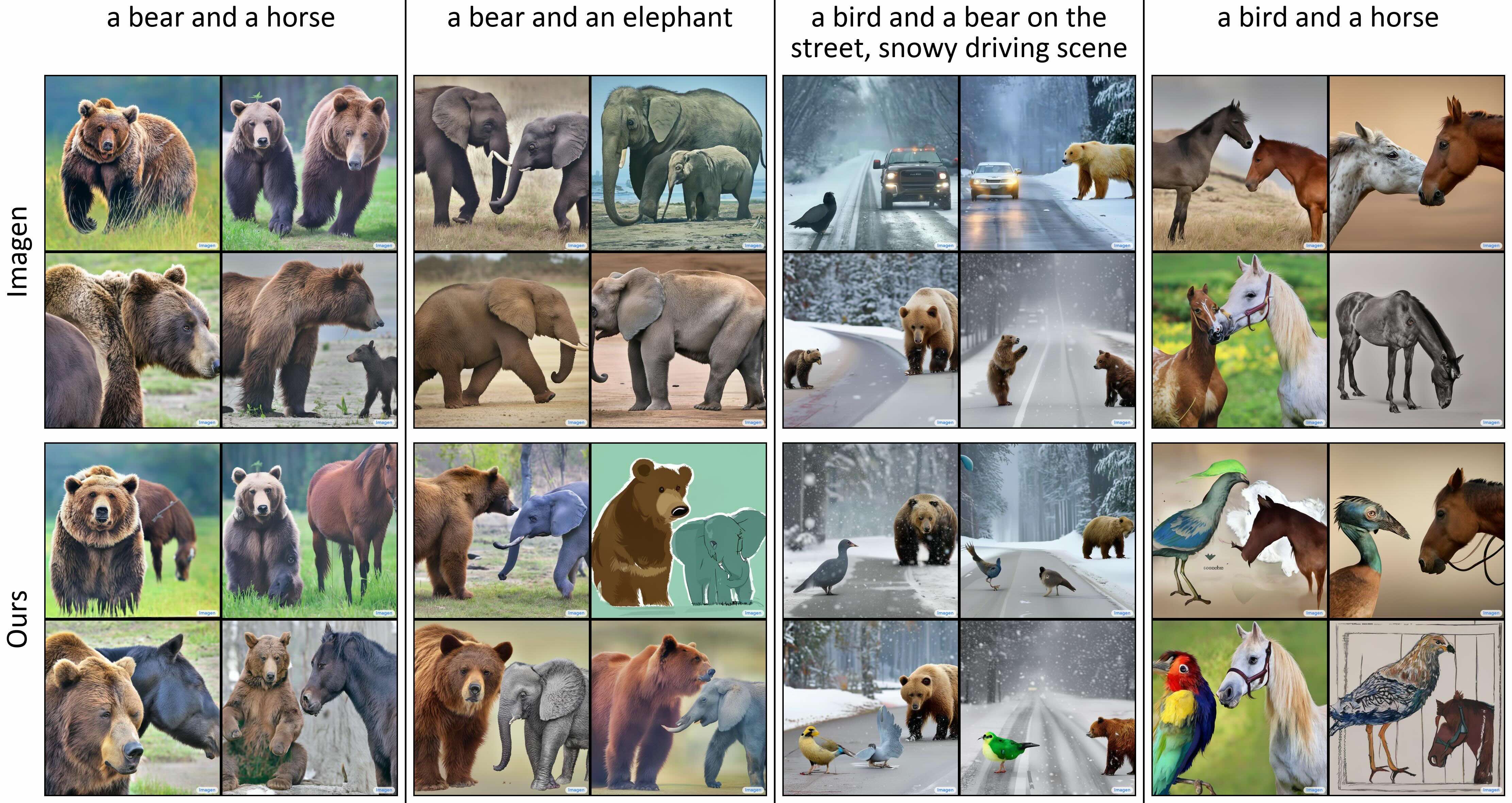}
    \caption{\textbf{Qualitative comparison of CONFORM on Imagen.} Our approach consistently produces images that more accurately reflect the input text prompts, effectively handling both simple and complex scenarios in the Imagen model. 
    }
    \label{fig:supplementary_imagen_page_0}
    \vspace{-1em}
\end{figure*}

\begin{figure*}[t!]
    \centering
    \includegraphics[width=0.89\linewidth]{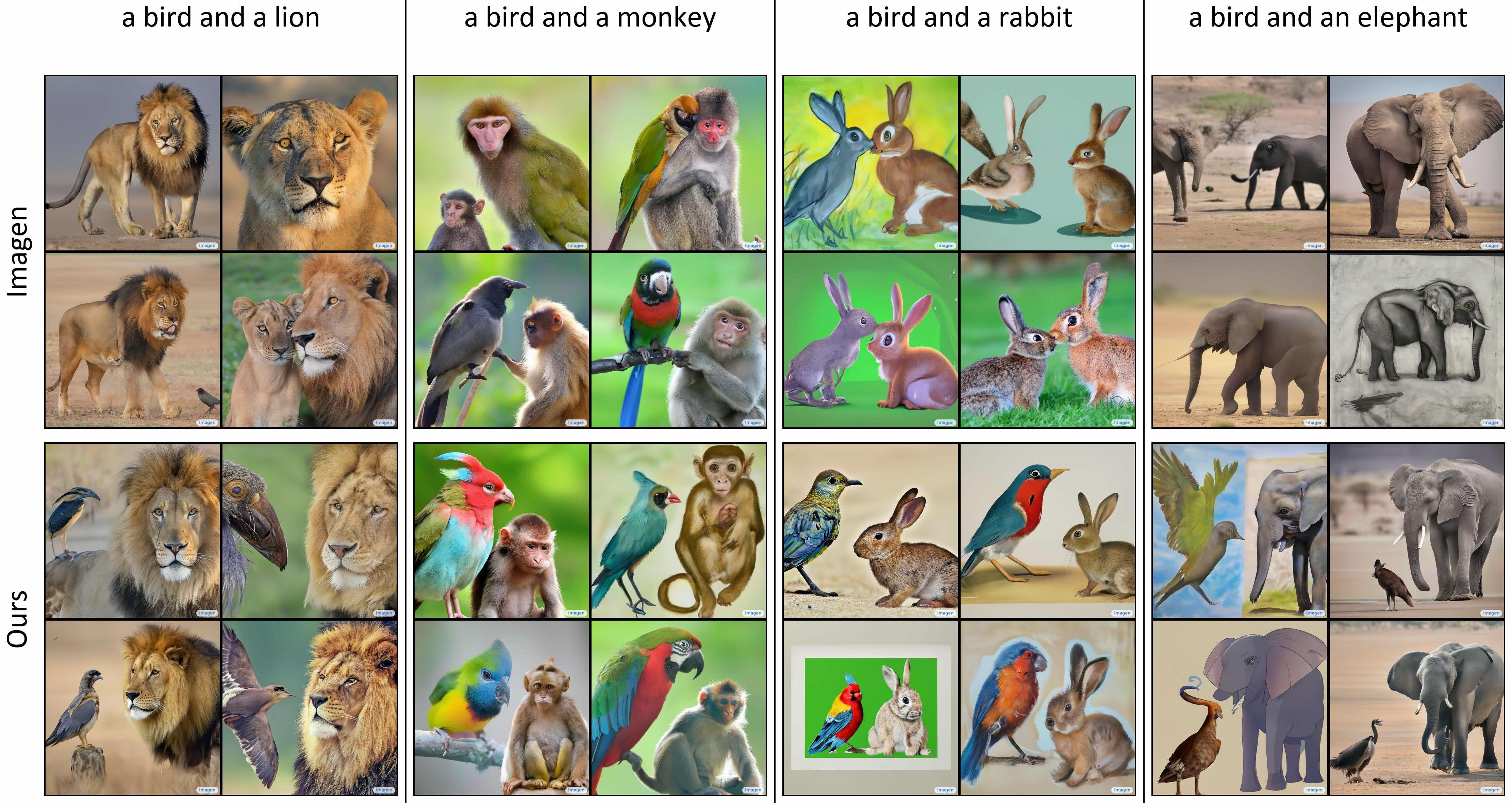}
    \caption{\textbf{Qualitative comparison of CONFORM on Imagen.} Our approach consistently produces images that more accurately reflect the input text prompts, effectively handling both simple and complex scenarios in the Imagen model. 
    }
    \label{fig:supplementary_imagen_page_1}
    \vspace{-1em}
\end{figure*}

\begin{figure*}[t!]
    \centering
    \includegraphics[width=0.89\linewidth]{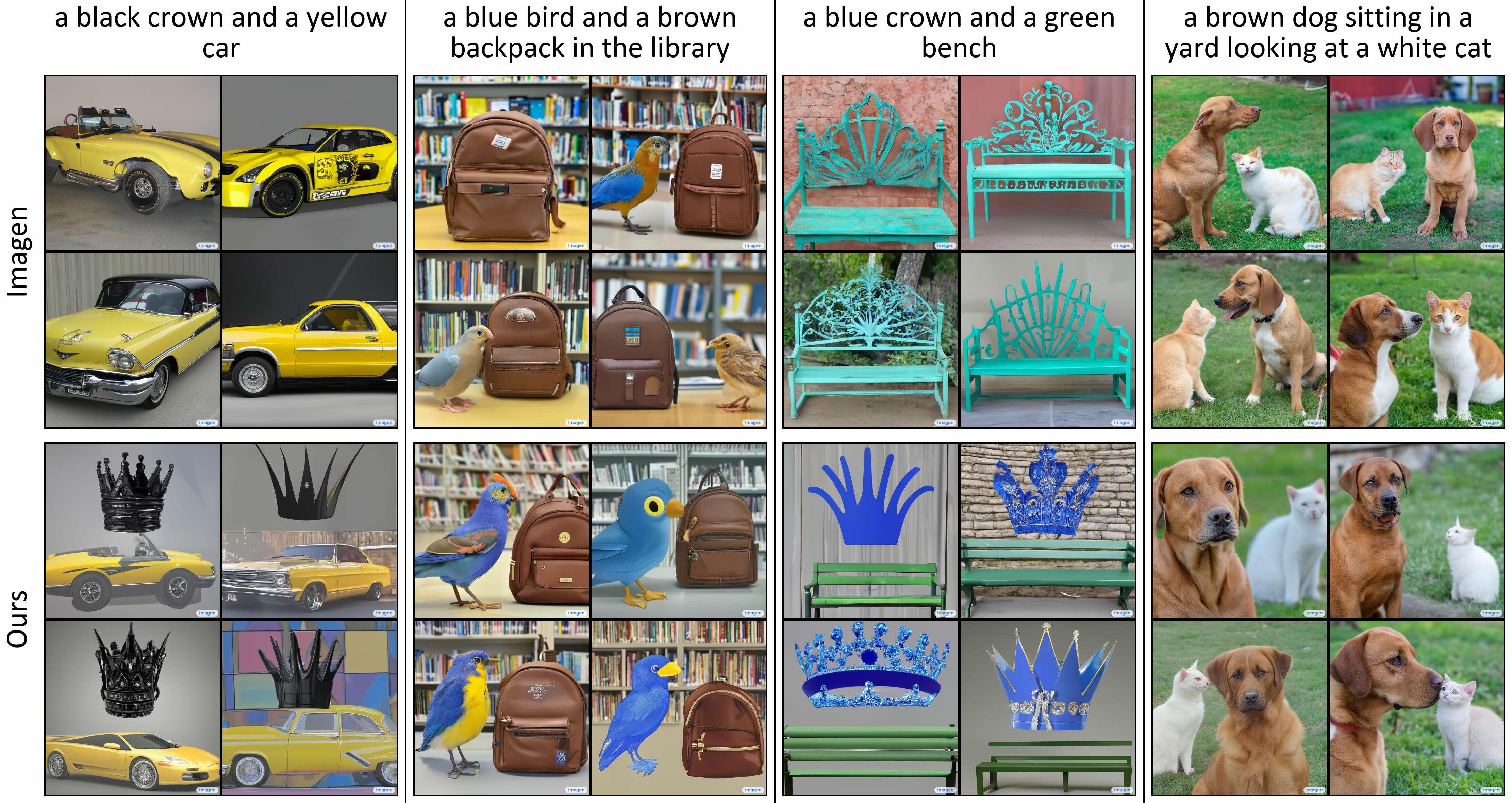}
    \caption{\textbf{Qualitative comparison of CONFORM on Imagen.} Our approach consistently produces images that more accurately reflect the input text prompts, effectively handling both simple and complex scenarios in the Imagen model. 
    }
    \label{fig:supplementary_imagen_page_2}
    \vspace{-1em}
\end{figure*}

\begin{figure*}[t!]
    \centering
    \includegraphics[width=0.89\linewidth]{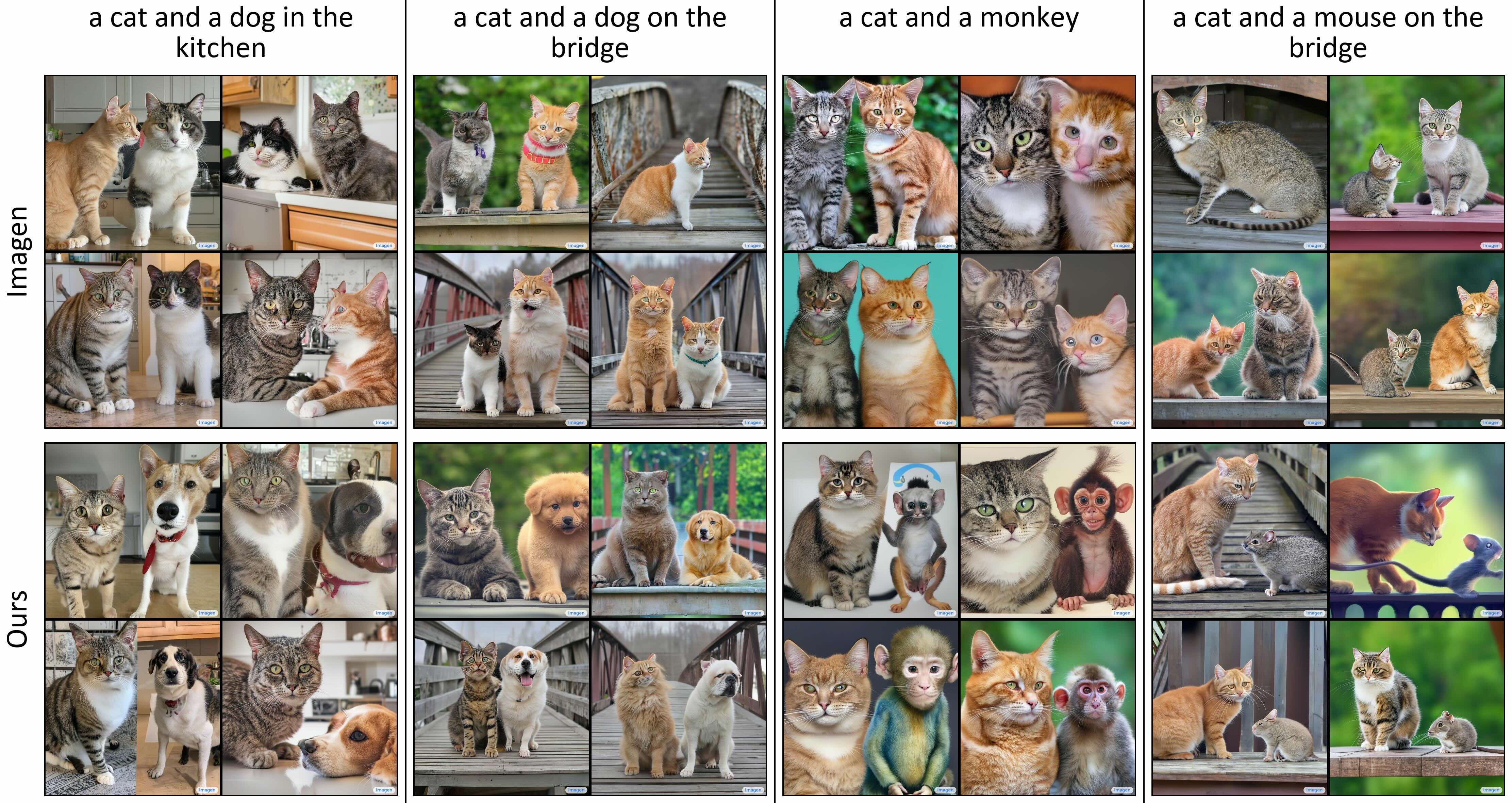}
    \caption{\textbf{Qualitative comparison of CONFORM on Imagen.} Our approach consistently produces images that more accurately reflect the input text prompts, effectively handling both simple and complex scenarios in the Imagen model. 
    }
    \label{fig:supplementary_imagen_page_3}
    \vspace{-1em}
\end{figure*}

\begin{figure*}[t!]
    \centering
    \includegraphics[width=0.89\linewidth]{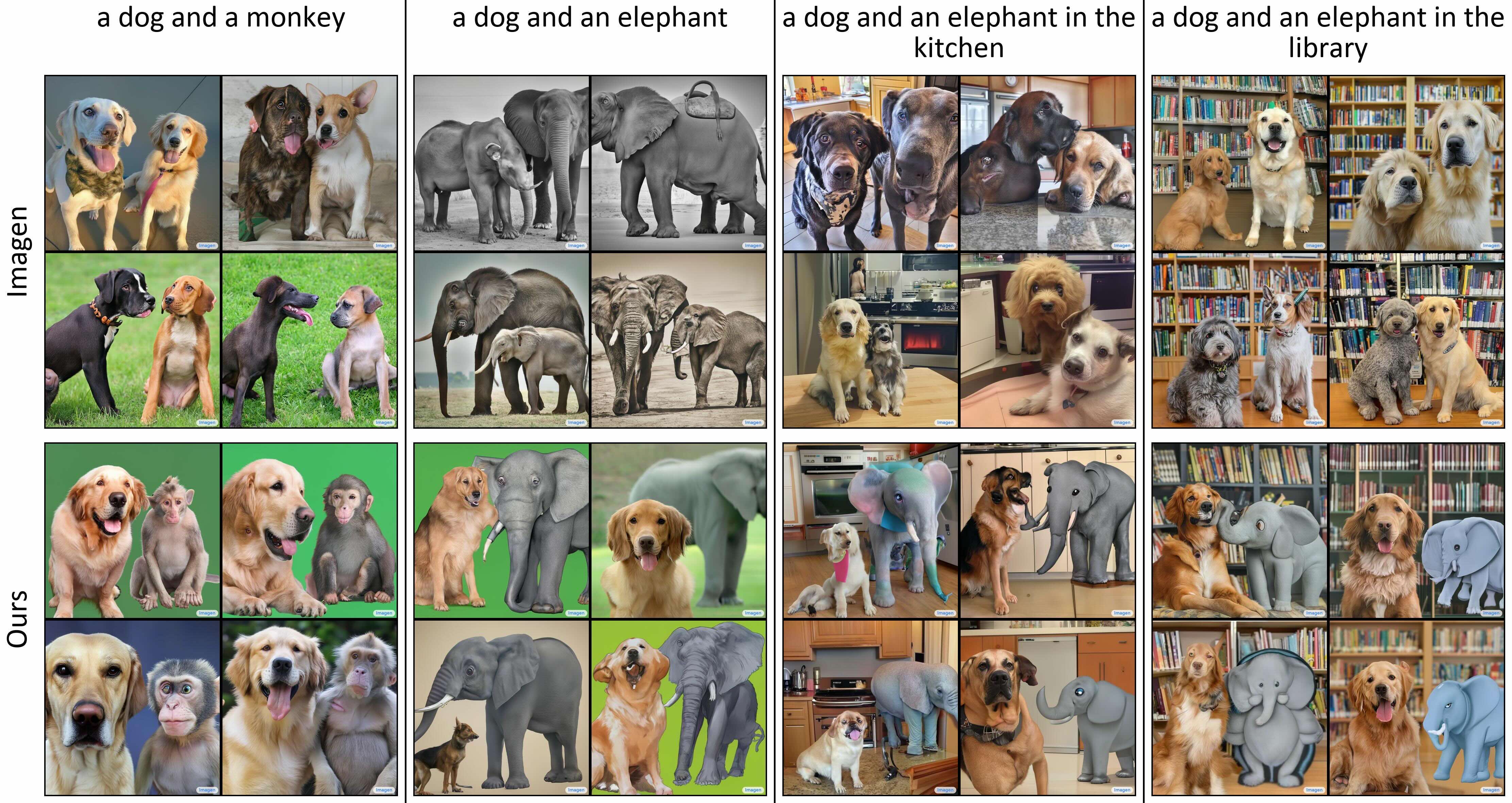}
    \caption{\textbf{Qualitative comparison of CONFORM on Imagen.} Our approach consistently produces images that more accurately reflect the input text prompts, effectively handling both simple and complex scenarios in the Imagen model. 
    }
    \label{fig:supplementary_imagen_page_4}
    \vspace{-1em}
\end{figure*}

\begin{figure*}[t!]
    \centering
    \includegraphics[width=0.89\linewidth]{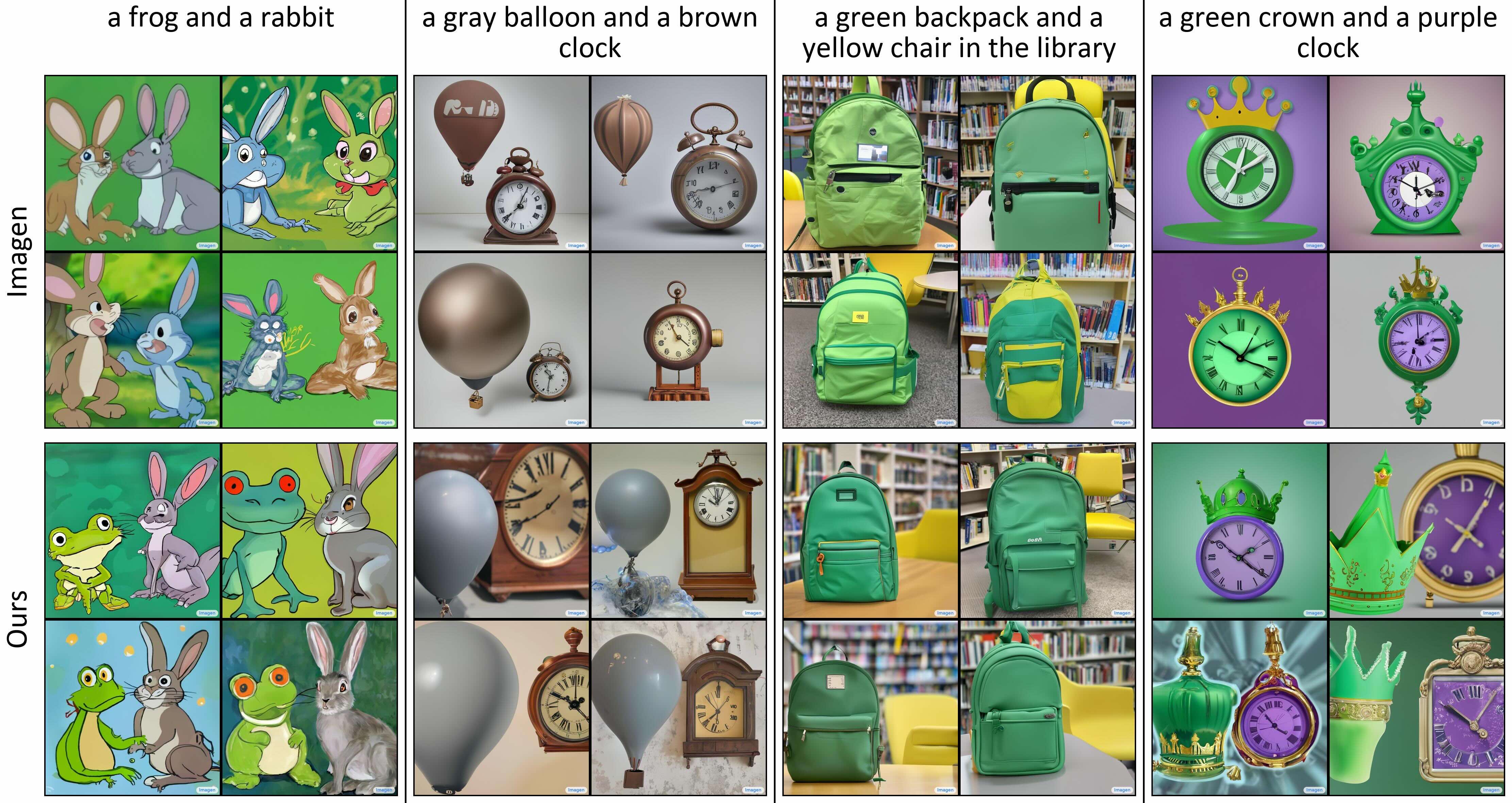}
    \caption{\textbf{Qualitative comparison of CONFORM on Imagen.} Our approach consistently produces images that more accurately reflect the input text prompts, effectively handling both simple and complex scenarios in the Imagen model. 
    }
    \label{fig:supplementary_imagen_page_5}
    \vspace{-1em}
\end{figure*}

\begin{figure*}[t!]
    \centering
    \includegraphics[width=0.89\linewidth]{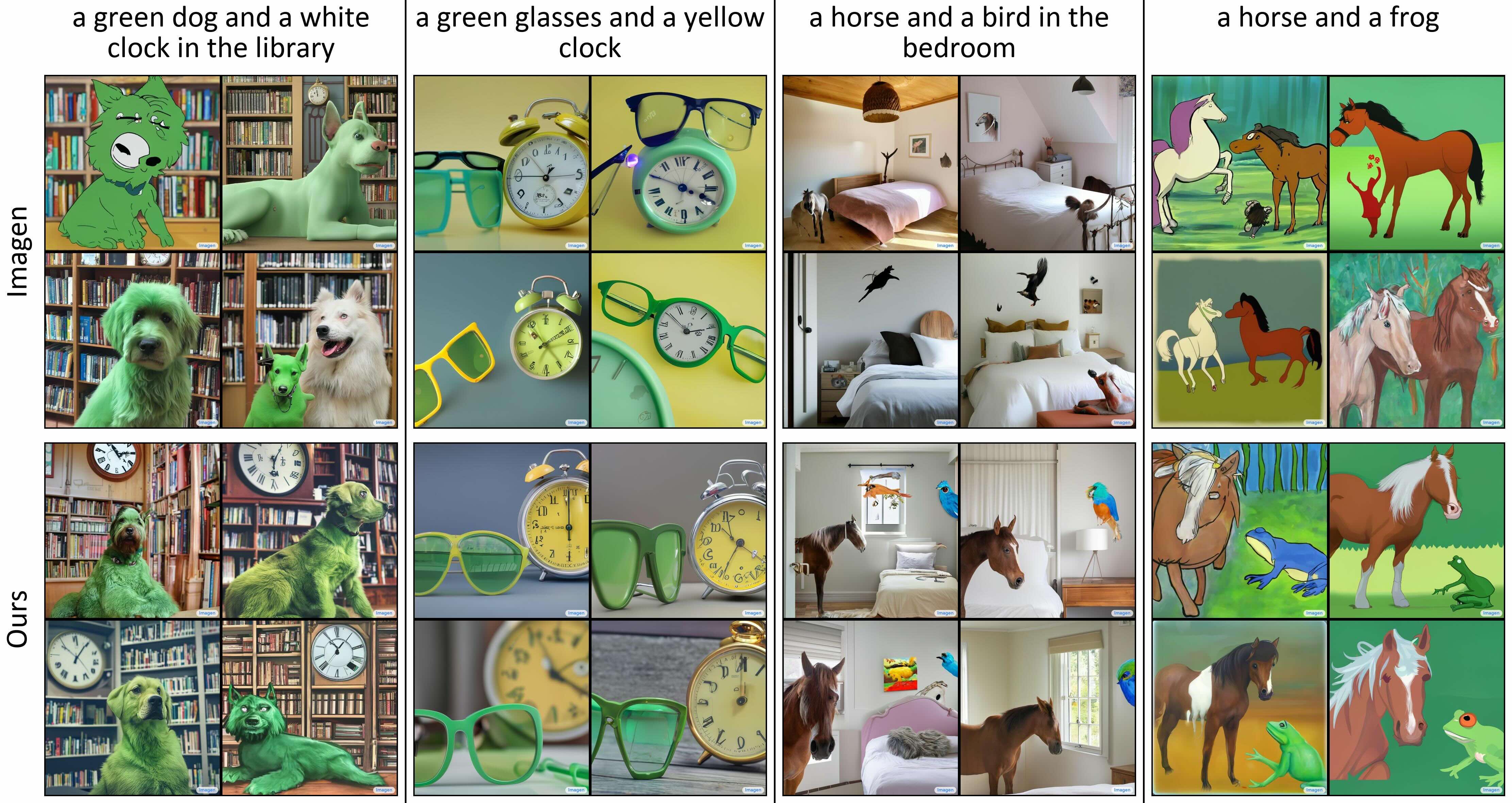}
    \caption{\textbf{Qualitative comparison of CONFORM on Imagen.} Our approach consistently produces images that more accurately reflect the input text prompts, effectively handling both simple and complex scenarios in the Imagen model. 
    }
    \label{fig:supplementary_imagen_page_6}
    \vspace{-1em}
\end{figure*}

\begin{figure*}[t!]
    \centering
    \includegraphics[width=0.89\linewidth]{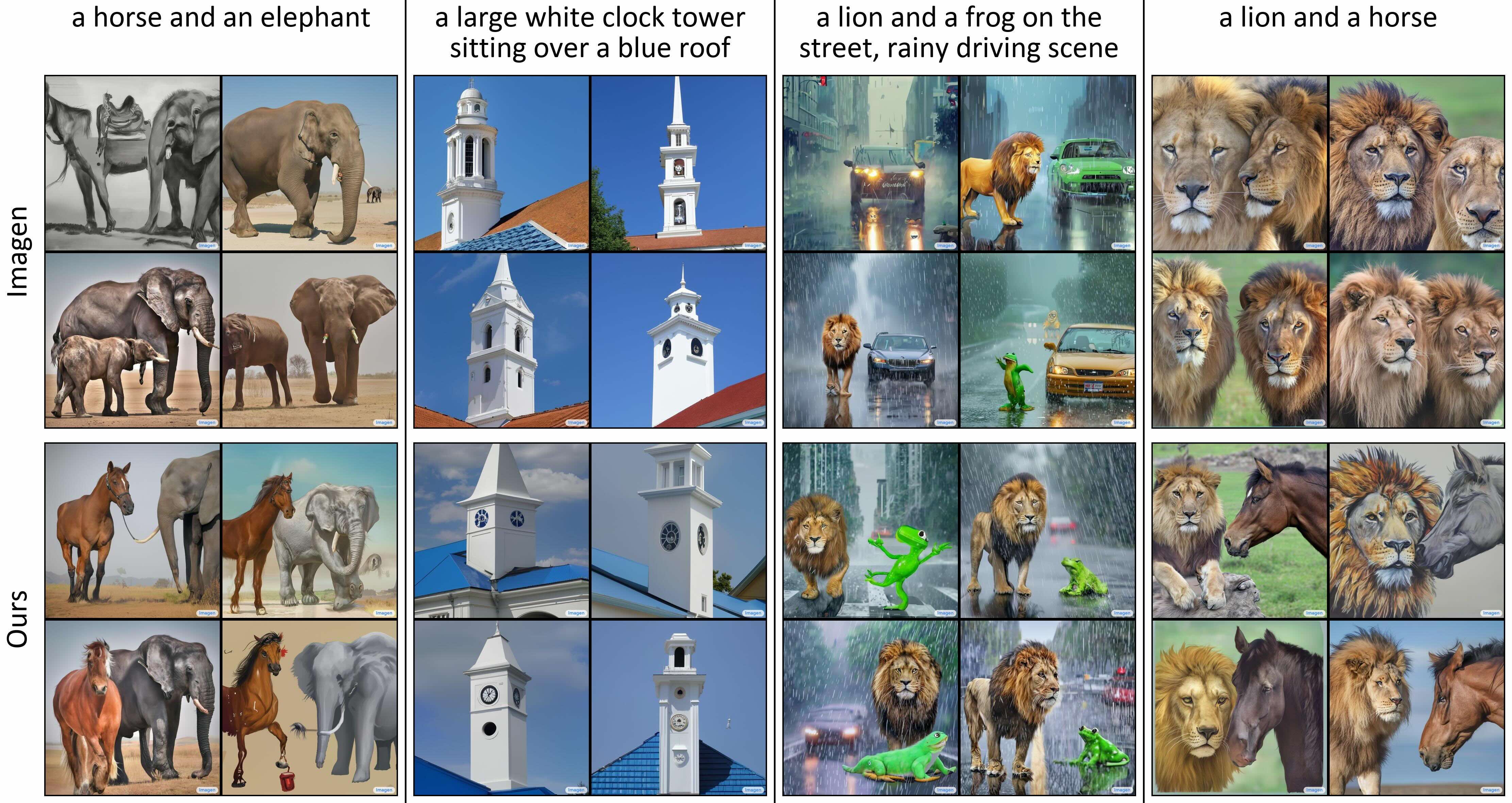}
    \caption{\textbf{Qualitative comparison of CONFORM on Imagen.} Our approach consistently produces images that more accurately reflect the input text prompts, effectively handling both simple and complex scenarios in the Imagen model. 
    }
    \label{fig:supplementary_imagen_page_7}
    \vspace{-1em}
\end{figure*}

\begin{figure*}[t!]
    \centering
    \includegraphics[width=0.89\linewidth]{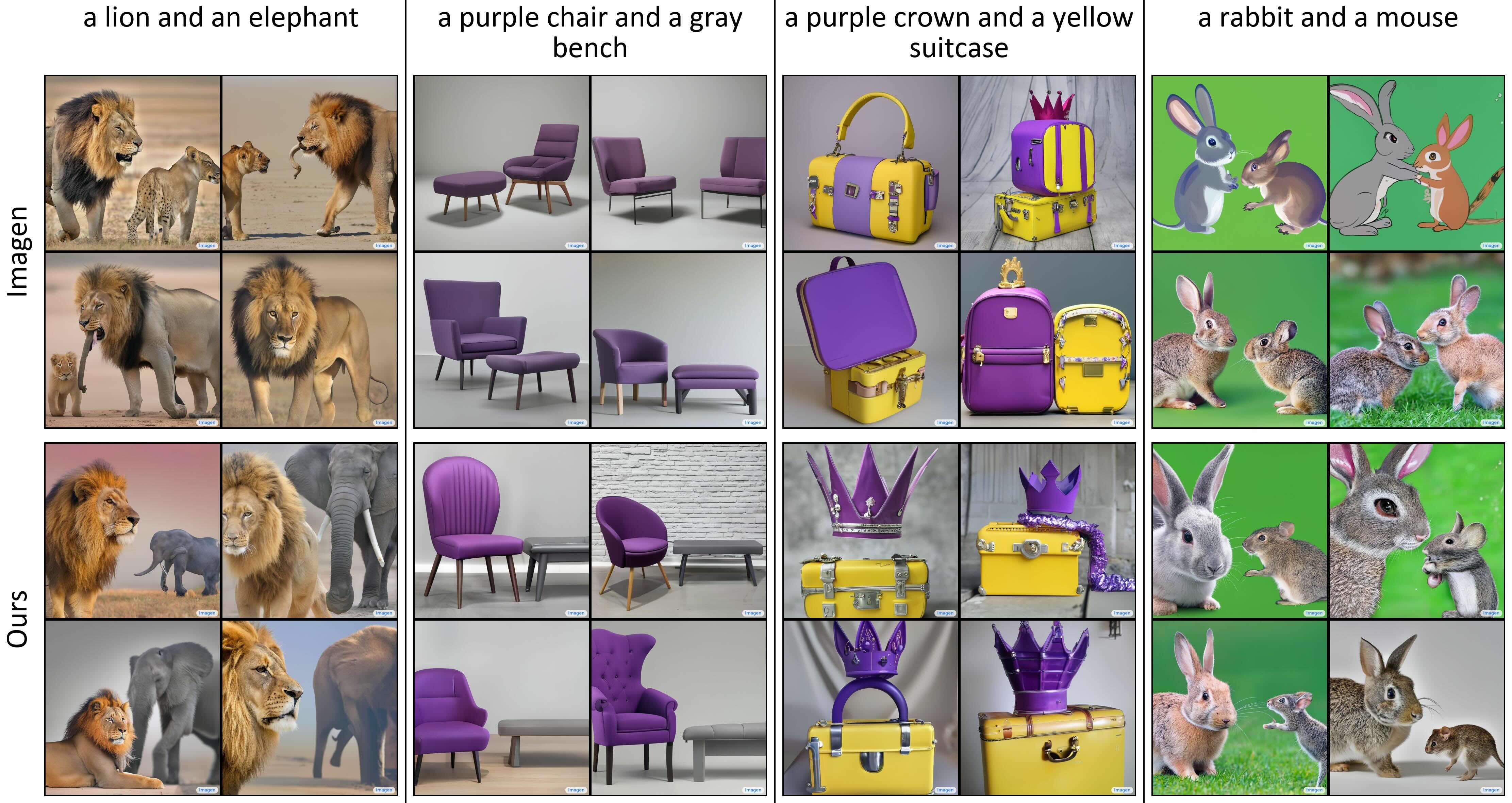}
    \caption{\textbf{Qualitative comparison of CONFORM on Imagen.} Our approach consistently produces images that more accurately reflect the input text prompts, effectively handling both simple and complex scenarios in the Imagen model. 
    }
    \label{fig:supplementary_imagen_page_8}
    \vspace{-1em}
\end{figure*}

\begin{figure*}[t!]
    \centering
    \includegraphics[width=0.89\linewidth]{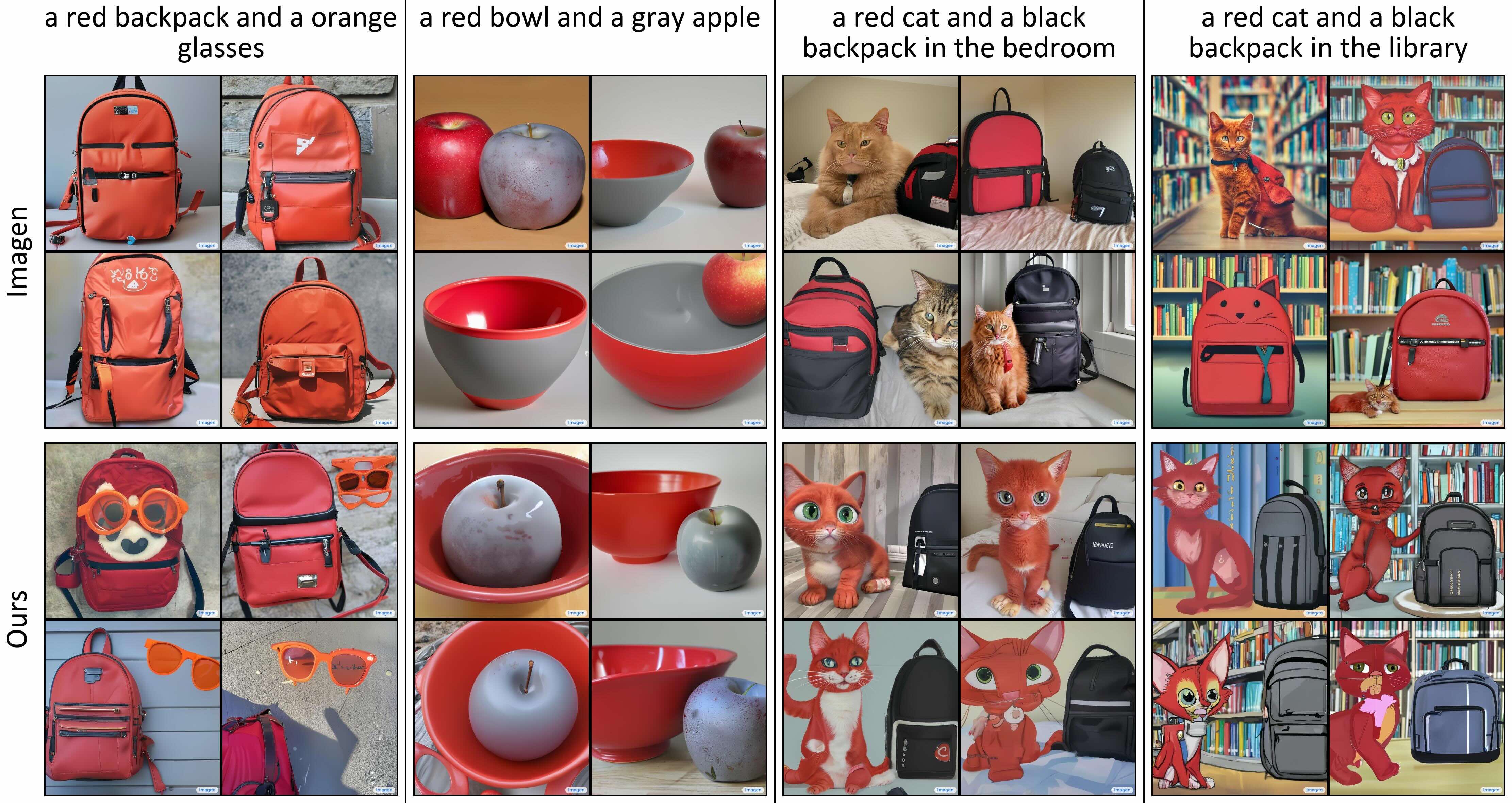}
    \caption{\textbf{Qualitative comparison of CONFORM on Imagen.} Our approach consistently produces images that more accurately reflect the input text prompts, effectively handling both simple and complex scenarios in the Imagen model. 
    }
    \label{fig:supplementary_imagen_page_9}
    \vspace{-1em}
\end{figure*}

\begin{figure*}[t!]
    \centering
    \includegraphics[width=0.89\linewidth]{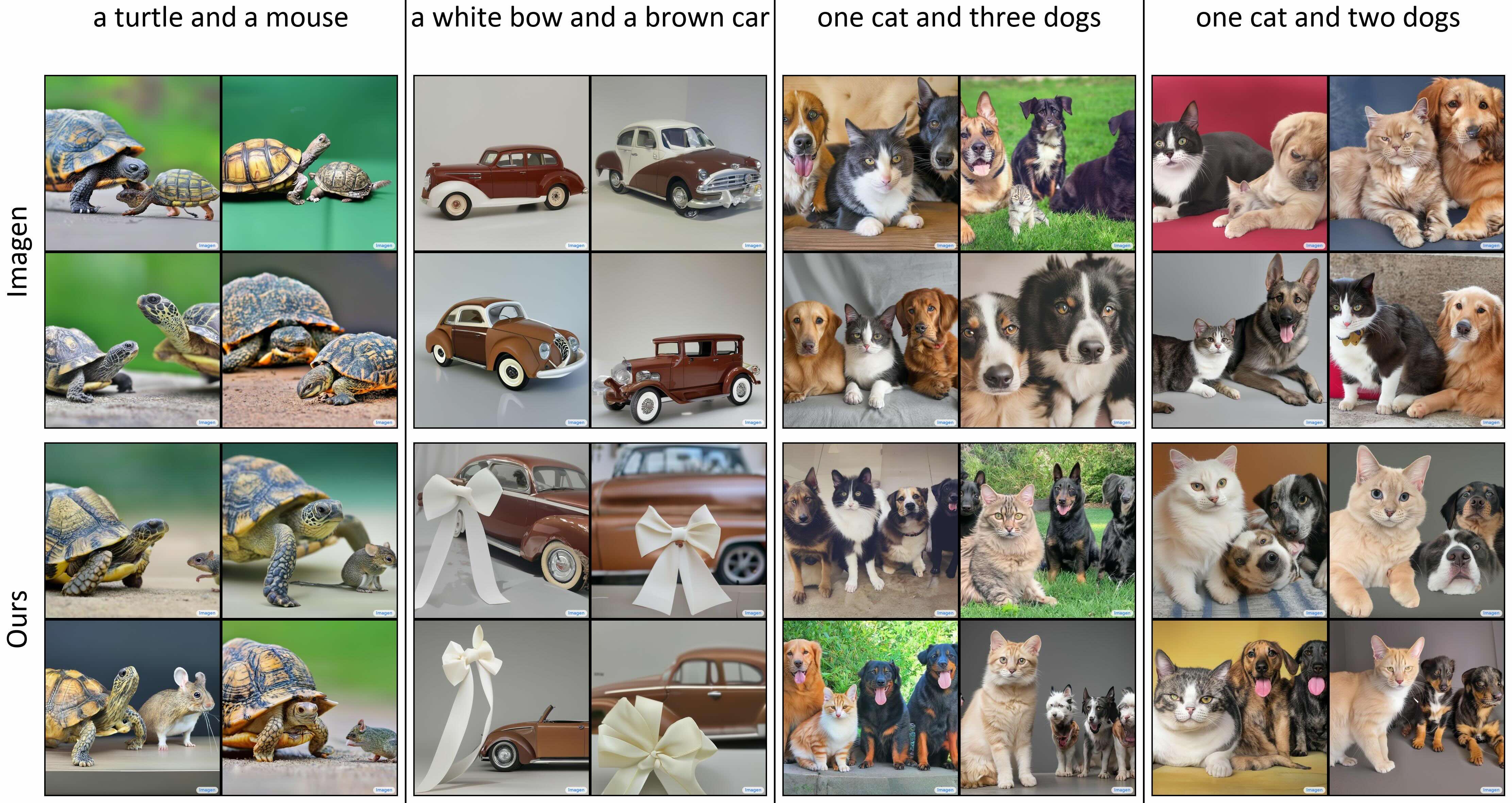}
    \caption{\textbf{Qualitative comparison of CONFORM on Imagen.} Our approach consistently produces images that more accurately reflect the input text prompts, effectively handling both simple and complex scenarios in the Imagen model. 
    }
    \label{fig:supplementary_imagen_page_10}
    \vspace{-1em}
\end{figure*}

\begin{figure*}[t!]
    \centering
    \includegraphics[width=0.89\linewidth]{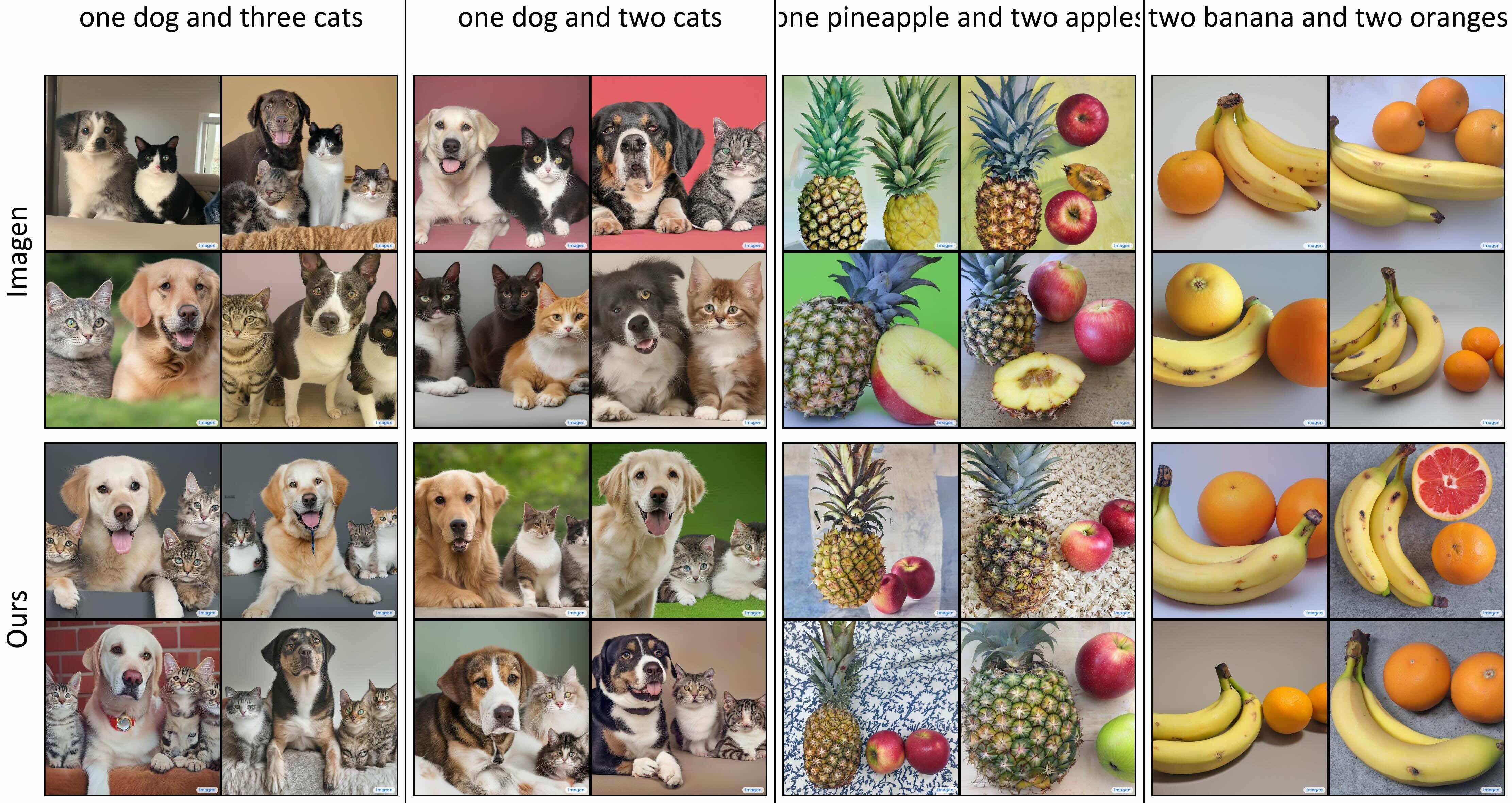}
    \caption{\textbf{Qualitative comparison of CONFORM on Imagen.} Our approach consistently produces images that more accurately reflect the input text prompts, effectively handling both simple and complex scenarios in the Imagen model. 
    }
    \label{fig:supplementary_imagen_page_11}
    \vspace{-1em}
\end{figure*}

\end{document}